ENHANCING TINYBERT FOR FINANCIAL SENTIMENT ANALYSIS USING GPT-AUGMENTED FINBERT DISTILLATION

GRAISON JOS THOMAS

Final Thesis Report

JUNE 2024

**Table of Contents**

# ABSTRACT

In the rapidly evolving field of financial sentiment analysis, the efficiency and accuracy of predictive models are critical due to their significant impact on financial markets. Transformer based models such as BERT, and more recently, large language models (LLMs) like GPT-4, have advanced NLP tasks considerably. Despite their advantages, BERT-based models face challenges with computational intensity in edge computing environments, and the substantial size and compute requirements of LLMs limit their practical deployment.

This study proposes leveraging the generative capabilities of LLMs, such as GPT-4 Omni, to create synthetic, domain-specific training data. This approach addresses the challenge of data scarcity and enhances the performance of smaller models by making them competitive with their larger counterparts. The research specifically aims to enhance FinBERT, a BERT model fine-tuned for financial sentiment analysis, and develop TinyFinBERT, a compact transformer model, through a structured, two-tiered knowledge distillation strategy.

Using data augmented by GPT-4 Omni, which involves generating new training examples and transforming existing data, we significantly improved the accuracy of FinBERT, preparing it to serve as a teacher model. This enhanced FinBERT then distilled knowledge to TinyFinBERT, employing both GPT-4 Omni and GPT-3.5 Turbo augmented data. The distillation strategy incorporated both logit and intermediate layer distillation. The training and evaluation of TinyFinBERT utilized the PhraseBank dataset and the FiQA 2018 Task1 dataset, achieving performance comparable to FinBERT while being substantially smaller and more efficient.

This research demonstrates how LLMs can effectively contribute to the advancement of financial sentiment analysis by enhancing the capabilities of smaller, more efficient models through innovative data augmentation and distillation techniques.

# LIST OF ABBREVIATION

| | |
|---|---|
| **AUC** | Area under the curve |
| **BERT** | Bidirectional Encoder Representations from Transformers |
| **CNN** | Convolutional Neural Networks |
| **DA** | Data Augmentation |
| **GPT** | Generative Pre-trained Transformer |
| **KD** | Knowledge Distillation |
| **LLM** | Large Language Model |
| **NLP** | Natural Language Processing |
| **RNN** | Recurrent Neural Networks |

**List of Figures**

# LIST OF TABLES

# CHAPTER 1: INTRODUCTION

## 1.1 Background

The rapid evolution of digital technology has profoundly impacted financial markets, leading to an era marked by a significant reliance on Internet-based platforms such as social media, blogs, digital news, and specialized financial forums like Twitter and Stocktwits (Agarwal, 2023; Nyakurukwa and Seetharam, 2023). These platforms have become pivotal in shaping public perceptions and market trends through the vast dissemination and exchange of opinions, comments, and reviews. This transformation has given rise to a rich repository of textual data, offering unprecedented opportunities for financial sentiment analysis (Sohangir et al., 2018; Agarwal, 2023). Leveraging such data, financial analysts, traders, and automated systems can now access real-time insights into market sentiments, potentially predicting market movements with greater accuracy than ever before.

Financial sentiment analysis, a key area of focus, involves extracting and interpreting emotions, opinions, and sentiments from textual data associated with financial markets, including news articles, social media posts, and financial reports. The primary goal of this analysis is to discern the prevailing sentiment towards financial assets or entities, such as stocks, currencies, and companies, which plays a crucial role in predictive analytics and investment decisions. By analysing market sentiment, stakeholders can make informed choices regarding asset allocation and stock trading, using these insights to proactively anticipate and respond to market dynamics. The nuanced understanding of market sentiment, facilitated by advances in machine learning and natural language processing (NLP) technologies, has become integral to strategic decision-making in finance (Sohangir et al., 2018; Messina et al., 2020; Joiner et al., 2022).

In recent years, the field of NLP has seen substantial advancements due to the introduction of machine learning technologies and transformer-based models like BERT (Bidirectional Encoder Representations from Transformers) (Devlin et al., 2019). These technologies have set new benchmarks in understanding the nuances of human language and enhanced the performance of various downstream NLP tasks through innovative approaches such as sequential transfer learning. This 'pretrain then fine-tune' paradigm, which leverages the generic knowledge acquired during pretraining on vast corpora (Chan et al., 2023), has been particularly effective in domains like finance, where models like FinBERT (Araci, 2019; Huang et al., 2023; Liu et al., 2021; Y. Yang et al., 2020) have been specifically fine-tuned to achieve superior performance in financial sentiment analysis.

In recent years, large language models (LLMs) have demonstrated remarkable capabilities in various natural language processing (NLP) tasks, including financial sentiment analysis. Models such as GPT-3 (Brown et al., 2020), GPT-4 Omni (Achiam et al., 2023), and other advanced LLMs have showcased impressive few-shot learning abilities, enabling them to perform well even with limited task-specific data (Hoffmann et al., 2022; Thoppilan et al., 2022; Zhang et al., 2022; Chowdhery et al., 2023). These capabilities are built upon the Transformer architecture (Vaswani et al., 2017) and large-scale textual corpora, leveraging transfer learning for enhanced performance (Radford et al., 2018, 2019; Devlin et al., 2019).

More recently as per (Li et al., 2023a) large language models (LLMs), renowned for their capacity to understand and generate human-like text, are also being harnessed to tackle complex financial analysis tasks, including financial sentiment analysis, thereby offering innovative solutions and insights. The approaches to deploying LLMs in finance have varied, ranging from the utilization of pretrained models such as LLaMA (Touvron et al., 2023b; a), BLOOM (Workshop et al., 2022), and Flan-T5 (Chung et al., 2022), through zero-shot or few-shot learning methods. These methods allow the models to perform specialized tasks without extensive, task-specific training. Another approach involves fine-tuning these LLMs on domain-specific data to create finely-tuned finance LLMs like PIXIU (Xie et al., 2023), FinGPT (Yang et al., 2023), and Instruct-FinGPT (Zhang et al., 2023). In addition to leveraging existing LLMs, there is a growing trend of developing custom LLMs from scratch to meet the unique needs of the finance sector. Models like BloombergGPT (Wu et al., 2023) and Fin-T5 (Lu et al., 2023) represent this innovative direction, offering solutions crafted explicitly for financial data analysis, forecasting, and reporting.

As per Hsieh et al., 2023, despite their potential, the deployment of LLMs faces significant hurdles due to their enormous size and computational demands, often requiring specialized infrastructure that limits their use in real-time applications. To overcome these challenges, practitioners have turned to smaller, specialized models that are more suited for real-time applications. These models are primarily trained using two methods: fine-tuning and knowledge distillation. Fine-tuning updates a pre-trained model, such as the Bidirectional Encoder Representations from Transformers (BERT, Devlin et al., 2019) or the Text-to-Text Transfer Transformer (T5, Raffel et al., 2020), with domain-specific, human-annotated data (Howard and Ruder, 2018a). Alternatively, knowledge distillation involves training smaller models using labels generated by a larger LLM (Tang et al., 2019a; Raffel et al., 2020; Wang et al., 2021; Smith et al., 2024). While both methods effectively reduce model size, they introduce their own challenges: fine-tuning requires extensive and expensive human-labelled data, particularly challenging in specialized domains like finance, and knowledge distillation demands large amounts of unlabelled data, which can also be difficult to procure (Tang et al., 2019a; Liang et al., 2020).

However, the computational complexity and resource requirements of such state-of-the-art LLMs and transformer models like BERT pose significant challenges, especially in real-time analysis and applications constrained by computational resources (Jiao et al., 2019; Gou et al., 2021). Prior research (Kovaleva et al., 2019; Michel et al., 2019; Voita et al., 2019) has revealed the presence of redundancy within Pre-trained Language Models (PLMs). As per (Kovaleva et al., 2019), complex models including fine-tuned BERT variants, exhibit over-parametrization when applied to domain-specific tasks. Consequently, this suggests that smaller models have the potential to match the performance levels of larger BERT models in such tasks (Jiao et al., 2019).

The current landscape of NLP in finance involves a blend of traditional methods and cutting-edge approaches, including the use of pretrained models for zero-shot or few-shot learning and the fine-tuning of these models on domain-specific data to enhance their applicability in finance. Despite these innovations, significant challenges remain, particularly in terms of the computational demands and the need for extensive, high-quality training data.

To address these challenges, this research aims to leverage the potential of large language models like GPT-4 Omni and GPT-3.5 Turbo for generating domain-specific training data and employing knowledge distillation techniques to enhance the performance of smaller, more computationally efficient models like FinBERT and TinyBERT. In this study a two-tiered knowledge distillation strategy: initially enhancing FinBERT using GPT-4 Omni-augmented data and subsequently distilling this knowledge to TinyBERT. This approach aims to develop computationally efficient models that retain high accuracy and robust performance in financial sentiment analysis, thus facilitating real-time applications in resource-constrained environments. This approach not only aims to mitigate the limitations posed by the size and complexity of traditional LLMs but also strives to harness their sophisticated capabilities for enhancing financial sentiment analysis.

By addressing the limitations of current paradigms and demonstrating the efficacy of LLM-augmented data for knowledge distillation, this research contributes to the advancement of efficient NLP models for financial sentiment analysis. The findings of this study have the potential to bridge the gap in current methodologies, offering a viable solution for deploying high-performance models in practical, real-world settings.

Consequently, there is a growing emphasis on developing distilled versions of these models to address such challenges. Several distilled versions of BERT, including DistilBERT (Sanh et al., 2019), TinyBERT (Jiao et al., 2019), and MobileBERT (Sun et al., 2020), have been developed specifically to mitigate the original model's computational and memory requirements.
TinyBERT, was introduced as a compact, efficient transformer model that promises to retain the sophisticated capabilities of its predecessors while drastically reducing the computational overhead. Despite its advantages, the performance of TinyBERT, particularly in specialized domains like finance, can be limited by its reduced size and the general nature of its pre-training. FinBERT (Araci, 2019), a variant of BERT fine-tuned specifically for financial sentiment analysis, demonstrates superior performance in this domain but suffers from the same computational inefficiencies that plague larger models.

This paper proposes a novel approach to enhance TinyBERT's performance in financial sentiment analysis by utilizing knowledge distillation techniques, leveraging the domain-specific strengths of FinBERT. Knowledge distillation is a technique where a smaller, simpler model (student) is trained to replicate the behavior of a larger, more complex model (teacher) to achieve comparable performance with reduced computational requirements. (Gou et al., 2021).

Moreover, recognizing the limitations posed by the availability of sufficient and diverse annotated financial datasets for training, this study integrates advanced data augmentation techniques powered by GPT-4 Omni (Achiam et al., 2023). This not only enriches the training data but also ensures the model's robustness and adaptability to new, unseen financial texts. By employing a combination of Soft Targets Distillation, Data Augmentation, and Layer-wise Distillation, and evaluating the models using the PhraseBank (Malo et al., 2014) and Forex News Annotated (Fatouros et al., 2023) datasets, this research aims to create a more efficient and accurate model for financial sentiment analysis.

The ensuing sections will detail the methodology employed in enhancing TinyBERT using GPT-augmented FinBERT distillation, the datasets used for training and evaluation, and the potential implications of these results in the development of efficient, real-time models for financial sentiment analysis, particularly advantageous in settings with limited resources.

## 1.2 Problem statement

In the field of financial sentiment analysis, the demand for efficient and accurate predictive models is crucial due to their substantial impact on financial markets. Transformer-based models like BERT, while advanced, encounter computational challenges in edge computing settings. Large language models (LLMs) such as GPT-4 Omni, though superior in performance, are hindered by their significant size and computational demands, limiting their practical deployment in real-time applications. Furthermore, methods like fine-tuning and knowledge distillation, essential for adapting these models to specific tasks, depend heavily on the availability of extensive, high-quality training data, which remains scarce in specialized domains like finance.

Recognizing the potential of LLMs in data augmentation for NLP tasks, there have been explorations into their application within the financial domain. However, no prior study, to the best of our knowledge, has specifically investigated the use of LLMs for data augmentation in financial sentiment analysis aimed at both fine-tuning and knowledge distillation. This thesis addresses this gap by employing the generative capabilities of LLMs to create synthetic, domain-specific training data, which not only tackles data scarcity but also enhances the training effectiveness of smaller, more deployable models. Specifically, this study leverages the capabilities of GPT-4 Omni and GPT-3.5 Turbo in a two-tiered knowledge distillation strategy to improve the performance of FinBERT and TinyBERT in financial sentiment analysis. By utilizing data augmented by GPT-4 Omni, including generating new examples and transforming existing data, this approach aims to refine these models' accuracy, enabling them to perform comparably to larger models while maintaining a smaller, more practical size.

**Research Gaps**
Despite significant advancements, several gaps remain unaddressed:

1. **Effectiveness of Knowledge Distillation using GPT-4 Omni for financial sentiment analysis**
   Limited research has been conducted on the effectiveness of knowledge distillation from a LLM like GPT-4 Omni on the performance of FinBERT, particularly in financial sentiment analysis. Assessing GPT-4 Omni's ability to enhance FinBERT's performance using the Financial PhraseBank could provide valuable insights into the applicability of knowledge distillation across similar financial domains.

2. **Impact of GPT-4 Omni-Augmented Data on generalization capabilities**

    The impact of using GPT-4 Omni to generate and label financial data on the generalization capabilities of FinBERT in diverse financial contexts remains unexplored. Investigating whether GPT-4 Omni-augmented data can substantially improve FinBERT's performance on an unseen dataset, such as the Forex News Annotated Dataset, is crucial, particularly in scenarios where labelled training data is scarce.

3. **Impact of using GPT-4 Omni on performance of Distilled Models**

    The development of smaller, efficient models like TinyBERT through knowledge distillation holds promise for computational efficiency without significant loss of performance. Conducting a comparative analysis of TinyBERT, created by leveraging LLMs, against a baseline FinBERT can help determine if LLMs can effectively support the creation of compact models for financial sentiment analysis that maintain competitive performance in diverse financial contexts.

Addressing these gaps, this research aims to evaluate the potential of GPT-4 Omni augmented data and knowledge distillation techniques to enhance the performance and generalizability of financial sentiment analysis models. Understanding these impacts can lead to the development of more efficient and effective models, making financial sentiment analysis more accessible and accurate across various applications.

## 1.3 Aim & Objectives

**Aim**

This study aims to leverage the advanced capabilities of generative large language models (LLMs), specifically GPT-4 Omni and GPT-3.5 Turbo, to enhance the training process of TinyFinBERT, a distilled, compact version of FinBERT optimized for financial sentiment analysis. By generating synthetic, domain-specific training data, this approach addresses the challenges of data scarcity and enriches the PhraseBank dataset, enabling effective knowledge distillation. The primary objective is to maintain the high accuracy and domain-specific expertise of the original FinBERT while reducing computational demands, thus making real-time analysis feasible in resource-constrained environments. Ultimately, this research seeks to validate a structured, two-tiered distillation strategy that utilizes LLM augmented data to train TinyFinBERT, achieving comparable performance with FinBERT but with significantly lower resource usage.

**Objectives**

To achieve the aforementioned aim, the following objectives have been set:

1. **To Optimize FinBERT's Training using GPT-4 Omni Augmented Data:**
    - Generate synthetic labelled data by using GPT-4 Omni to generate new training examples using appropriate prompting strategy.
    - Generate synthetic labelled data by using GPT-4 Omni to create variations of the train data in the Financial PhraseBank dataset.
    - Fine-tune FinBERT with synthetic, domain-specific training data generated by GPT-4 to create an enhanced model, Augmented FinBERT.
    - Quantitatively assess the improvement in FinBERT's performance metrics: accuracy, precision, recall, and F1 scores, after training with augmented data, using the test data set aside from the Financial PhraseBank dataset.

2. **To Test Augmented FinBERT's Generalization on Unseen Financial Data:**
    - Compare performance metrics for Augmented FinBERT with those of baseline FinBERT on FiQA 2018 Task1 dataset and the Forex News Annotated dataset. These

datasets were not used in training FinBERT or Augmented FinBERT and hence can be used to assess the generalization capabilities of these 2 models.

3. **To Implement Advanced Distillation from Augmented FinBERT to TinyFinBERT using unlabelled and labelled data created by GPT-3.5 Turbo and GPT-4 Omni:**
    - Generate synthetic unlabelled data by using GPT-3.5 Turbo to generate new training examples using appropriate prompting strategy.
    - Generate synthetic unlabelled data by using GPT-3.5 Turbo to create variations of the train data in the Financial PhraseBank dataset.
    - Perform knowledge distillation techniques to train TinyFinBERT.
    - Utilize synthetic unlabelled data generated by GPT-3.5 Turbo along with the earlier synthetic labelled data generated by GPT-4 Omni during knowledge distillation.
    - Utilize logit and intermediate layer representations derived from Augmented FinBERT on labelled and unlabelled data to train TinyFinBERT.

4. **To Evaluate and Compare the Performance of TinyFinBERT:**
    - Measure and compare performance metrics for TinyFinBERT's against both Augmented and baseline FinBERT on the PhraseBank dataset, focusing on accuracy, precision, recall, and F1 scores.

5. **To Test TinyFinBERT's Generalization on Unseen Financial Data:**
    - Assess TinyFinBERT's ability to analyse financial sentiments on two unseen datasets, FiQA 2018 Task1 dataset and the Forex News Annotated dataset. These datasets were not used in training any of the models and hence can be used to assess the generalization capability of TinyFinBERT.
    - Compare performance metrics for TinyFinBERT with those of baseline FinBERT and Augmented FinBERT on FiQA 2018 Task1 dataset and the Forex News Annotated dataset.

## 1.4 Research Questions

This study aims to answer the following research questions:

1. How does knowledge distillation from GPT-4 Omni affect the performance (accuracy, F1 score, precision, and recall) of FinBERT in financial sentiment analysis on the Financial PhraseBank dataset compared to its performance without GPT-4 Omni's knowledge?
2. Does the integration of GPT-4 Omni and GPT 3.5 generated and labelled financial statements, used for knowledge distillation, enhance the generalized performance of FinBERT in classifying financial sentiments on the unseen Forex News Annotated dataset, as measured by accuracy, F1 score, precision, and recall, compared to the baseline FinBERT model?
3. Can TinyFinBERT, distilled from an augmented FinBERT model using GPT-4 Omni and GPT 3.5 augmented data, achieve comparable performance in terms of accuracy, F1 score, precision, and recall to the baseline FinBERT model in financial sentiment classification on the PhraseBank dataset?
4. Is TinyFinBERT, derived through knowledge distillation using GPT-4 Omni and GPT 3.5 augmented data, capable of achieving performance (accuracy, F1 score, precision, and recall) comparable to the baseline FinBERT model in generalizing across diverse financial contexts on the Forex News Annotated dataset?

## 1.5 Scope of the Study

This research focuses on the development and evaluation of TinyFinBERT, a compact, efficient model for financial sentiment analysis derived through a structured, two-tiered knowledge distillation process from an augmented version of FinBERT. The study employs two distinct datasets: the PhraseBank dataset, comprising financial news statements, and the Forex News Annotated dataset, which contains news headlines related to the foreign exchange markets. These datasets are strategically utilized to train and validate TinyFinBERT, and to independently assess its generalization capabilities across different types of financial data.

A significant aspect of this study involves the use of LLMs for creating synthetic training data and augmenting existing datasets. Specifically, the gpt-4o-2024-05-13 (GPT-4 Omni) model is employed for tasks requiring generative and labelling capabilities, while the gpt-3.5-turbo-0125 (GPT-3.5 Turbo) model is utilized for generating unlabelled data. The use of GPT-3.5 Turbo is particularly noted for its cost efficiency, being approximately 10 times cheaper than the GPT-4 Omni model, thus enabling substantial data augmentation without incurring high computational costs.

The selection of TinyFinBERT as the primary model for this study highlights its potential for deployment in computationally constrained environments, demonstrating a scalable distillation approach that could be applied to larger BERT-based models. The research meticulously examines knowledge distillation techniques, incorporating both logit and intermediate layer data from the augmented FinBERT. This innovative use of LLMs to enrich training datasets addresses the challenge of data scarcity in the specialized domain of finance.

By enhancing both the performance and the deployment efficiency of financial sentiment analysis models, this study aims to contribute significant insights to the fields of natural language processing (NLP) and financial analytics. It seeks to validate a novel distillation strategy that not only maintains high accuracy and specificity in financial sentiment classification but also ensures practical applicability of sophisticated AI models in real-world financial market settings.

## 1.6    Significance of the Study

The significance of this study extends across multiple dimensions within the field of financial sentiment analysis, a domain critical to the effective interpretation and prediction of market trends based on textual data. By advancing the capabilities of sentiment analysis models through the integration of LLMs like GPT-4 and GPT-3.5, with advanced knowledge distillation techniques, this study contributes significantly to both theoretical advancements and practical applications in data science.

**Theoretical Contributions:**

This study enhances the theoretical framework of knowledge distillation by demonstrating an effective two-tiered distillation strategy that leverages LLM generated synthetic data for model training. The incorporation of distillation techniques based on intermediate layer and logit data from an augmented FinBERT into a more compact TinyFinBERT model proves that deep semantic insights and contextual understanding can be preserved for financial domain data even with significantly reduced model size and computational load. This not only supports the development of more efficient NLP models but also provides a framework for future research in applying these techniques to other complex datasets in other specialized domains.

**Practical Implications:**

This study addresses the need within financial markets for both rapid and accurate sentiment analysis. By creating a model with comparable accuracy of its larger predecessors while operating within the constraints of real-time, resource-limited environments, the findings offer potential benefits to financial analysts and traders. These stakeholders can potentially apply the insights from the creation of TinyFinBERT to other use cases that require reduced latency in sentiment analysis.

**Advancements in Financial Sentiment Analysis:**

Specifically, in the realm of financial sentiment analysis, this study pioneers the use of LLM generated synthetic data to overcome the challenges of data scarcity and variability in financial texts. By enriching training datasets with high-quality, domain-specific examples, the research ensures that the distilled models are not only efficient but also robust and capable of generalizing across different financial contexts. This is particularly significant for industries where the rapid assessment of financial news and market sentiments can dictate investment decisions and strategic financial planning.

**Broader Impact:**

The outcomes of this research add to the growing body of work promoting more sustainable practices in model training and deployment. Financial institutions can leverage these insights to deploy advanced analytics tools without the prohibitive costs associated with large-scale computational resources, thereby democratizing access to state-of-the-art technology.

## 1.7   Structure of the Study

This thesis is organized into six chapters, each designed to systematically address the research questions posed and to showcase the methodologies employed, the analysis conducted, and the implications of the findings. The organization of the chapters is intended to provide a coherent flow from the conceptualization of the problem to the presentation of the research findings and recommendations.

**Chapter 2: Literature Review**

This chapter examines existing research on sentiment analysis, focusing specifically on financial sentiment analysis using BERT-based models and Large Language Models (LLMs) like GPT-3.5 and GPT-4. It explores the theoretical underpinnings of knowledge distillation and the use of synthetic data in training machine learning models. The review provides a critical analysis of current methodologies and highlights gaps that this study aims to address.

**Chapter 3: Research Methodology**

This chapter describes the comprehensive methodology used to conduct the research, including the design of the two-tiered knowledge distillation strategy and the creation of synthetic training data using GPT-4 and GPT-3.5. It details the selection of datasets, PhraseBank and Forex News Annotated, and outlines the procedures for model training, distillation, and evaluation.

**Chapter 4: Analysis**

This chapter presents a detailed analysis of the data augmented by GPT models and the subsequent knowledge distillation process. It examines how the augmented FinBERT performs as a teacher model and discusses the integration of intermediate layer and logit data into TinyFinBERT. This analysis seeks to uncover the nuanced impacts of these strategies on model performance and efficiency.

**Chapter 5: Results and Evaluation**

The findings from the empirical tests conducted are reported in this chapter. It evaluates the performance of TinyFinBERT compared to the augmented FinBERT and the baseline models in terms of accuracy, precision, recall, and F1 scores. This chapter also assesses the generalization capabilities of TinyFinBERT across different financial contexts using the Forex News Annotated dataset.

**Chapter 6: Conclusions and Recommendations**

The final chapter synthesizes the study's findings, discussing the implications for the field of financial sentiment analysis and the broader domain of NLP. It reflects on the efficiency and efficacy of the proposed distillation strategy and suggests practical applications for the developed models. Recommendations for future research are also provided, focusing on potential improvements and the exploration of other applications and datasets.

# CHAPTER 2 : LITERATURE REVIEW

## 2.1 Introduction

This chapter provides a comprehensive review of the relevant literature that underpins the theoretical and practical aspects of this study. The review begins with an exploration of sentiment analysis, an area of natural language processing (NLP) that has significantly evolved with the advent of deep learning technologies. Following this, the focus shifts to BERT (Bidirectional Encoder Representations from Transformers) and its variants, which represent a cornerstone in modern NLP research due to their revolutionary impact on language understanding tasks.

The discussion then extends to knowledge distillation strategies for BERT, which facilitate the compression and efficiency of these models without substantial loss in performance. This is particularly pertinent for deploying high-capacity models in resource-constrained environments. In parallel, the chapter explores data augmentation techniques, which are critical for enhancing the robustness and generalization of models by artificially expanding the training dataset.

Further, the integration of knowledge distillation with data augmentation using large language models (LLMs) is examined, showcasing how these methods synergize to refine the training of smaller, more efficient models. The challenges associated with deploying large language models are also discussed, highlighting the practical limitations and considerations necessary for their effective application in real-world settings.

## 2.2 Sentiment Analysis

The evolution of sentiment analysis has been profoundly influenced by the emergence of Large Language Models (LLMs) such as BERT (Bidirectional Encoder Representations from Transformers) (Devlin et al., 2019) and GPT (Generative Pre-trained Transformer) (Radford et al., 2018), along with their successors. Introduced by pioneering works like (Vaswani et al., 2017) and (Radford et al., 2019), these models represent a significant departure from traditional machine learning approaches. By leveraging deep learning technologies, including Convolutional Neural Networks (CNNs) and Recurrent Neural Networks (RNNs), the field initially saw considerable advancements. However, the advent of transformer-based architectures marked a watershed moment. Transformer based models like BERT have not only set new benchmarks for understanding context and nuance in textual data but have also substantially enhanced the accuracy, efficiency, and applicability of sentiment analysis. This shift has made sentiment analysis a more powerful tool for comprehensively understanding and interpreting human emotions and opinions expressed in text, illustrating a remarkable progression towards capturing complex emotional nuances with greater depth and precision. Financial sentiment analysis has emerged as a critical area within NLP, driven by the need for accurate and timely analysis of financial texts to inform investment and trading decisions. Traditional machine learning approaches have gradually been supplanted by deep learning methods, offering nuanced understanding of financial sentiment. The specificity of financial language, with its nuanced terminology and significant impact on decision-making, presents unique challenges that domain-specific models like FinBERT aim to address. (Araci, 2019)

## 2.3  BERT and Its Variants in NLP

The introduction of Bidirectional Encoder Representations from Transformers (BERT) by (Devlin et al., 2019) marked a turning point in NLP, setting new standards for a variety of tasks, including sentiment analysis, named entity recognition, and question answering. BERT employs an encoder-only architecture, distinguishing itself through a novel pre-training objective known as the "masked language model". This approach, which involves predicting randomly masked tokens within an input sequence, enables the model to capture deep, bidirectional representations of language. BERT-like models undergo two primary phases of training: pre-training and fine-tuning. Initially, these models are pre-trained on vast, general-purpose language datasets, allowing them to develop a comprehensive understanding of language structure and context. Subsequently, they are fine-tuned on smaller, task-specific datasets, a process that adapts their broad linguistic capabilities to particular applications. This methodology has propelled BERT and its successors to achieve state-of-the-art results across numerous NLP tasks.

In response to the unique challenges presented by financial language, domain-specific models like FinBERT (Araci, 2019) have been crafted. FinBERT represents a pivotal advancement in financial NLP by tailoring BERT specifically for financial sentiment analysis. The model underwent pre-training on the extensive TRC2-financial corpus, containing millions of words from finance-related Reuters news articles, to grasp finance-specific terminology effectively. Subsequently, FinBERT was fine-tuned with a classification layer on the Financial PhraseBank (Malo et al., 2014) dataset, significantly outperforming previous models in sentiment analysis accuracy on Financial PhraseBank and FiQA (Maia et al., 2018) datasets. This breakthrough addressed the critical need for specialized models capable of interpreting the complex nuances of financial language, which is vital for analysing market sentiments and guiding investment decisions.

Despite the emergence of various FinBERT models (Araci, 2019; Liu et al., 2021; Huang et al., 2023; Jiang and Zeng, 2023), a review of the current literature reveals a notable absence of efforts aimed at creating more compact and efficient versions of FinBERT tailored for financial sentiment analysis. This gap highlights an unexplored avenue for research dedicated to optimizing the balance between computational efficiency and performance accuracy in financial NLP.

## 2.4    Knowledge Distillation for BERT

Kovaleva et al., 2019, have indicated that fine-tuned BERT models are often over-parametrized for specific tasks, suggesting that smaller models could achieve comparable performance in these domains (Jiao et al., 2019). Knowledge distillation, introduced by (Hinton et al., 2015), is a popular technique for model compression. It aims to transfer knowledge from a large, more complex model (teacher) to a smaller, more efficient one (student), without significant loss in performance. This approach has gained popularity in NLP as a means to retain the sophisticated understanding capabilities of models like BERT in compact versions suitable for deployment in resource-constrained environments. DistilBERT (Sanh et al., 2019), the first BERT model compressed using this method, was followed by innovations like TinyBERT (Jiao et al., 2019), and MobileBERT (Sun et al., 2020) each bringing unique contributions to the process.

Task-specific distillation represents a targeted methodology within the wider field of knowledge distillation, aimed specifically at enhancing the performance of a smaller, student model on a particular task. This approach has been investigated in several studies (Sun et al., 2019; Tang et al., 2019b; Turc et al., 2019; Aguilar et al., 2020; Mukherjee and Hassan Awadallah, 2020; Xu et al., 2020), focusing on the efficient compression of large pre-trained language models into more compact forms. The process typically begins with fine-tuning these pre-trained language models on designated tasks. Following this fine-tuning phase, the distilled knowledge is then transferred to the smaller model, thereby optimizing its task-specific performance while maintaining a reduced model size.

In their study on TinyBERT, (Jiao et al., 2019) introduced a novel two-stage knowledge distillation approach for Transformer models. This method includes general distillation to imbue TinyBERT with the broad knowledge from pre-trained BERT, and task-specific distillation that uses an augmented dataset tailored to a specific task, allowing TinyBERT to acquire specialized knowledge from the BERT model that has been fine-tuned for that particular task. In this study, we will also adopt a similar strategy by utilizing a task-specific augmented dataset during the distillation process.

In the development of FinBERT, although the model underwent pre-training on the TRC2-financial corpus, this step did not significantly enhance its performance, as noted by Araci, 2019. Moreover, the TRC2 dataset is not publicly available. Therefore, this study concentrates solely on task-specific distillation to refine FinBERT's financial sentiment classification capabilities, excluding general distillation due to the limited efficacy of pre-training and the dataset's inaccessibility.

Beyond knowledge distillation, model compression also involves techniques like pruning, quantization, and structured sparsity, offering paths to reduce models' size and computational demands (Gou et al., 2021). However, these methods were not considered in this study, focusing our examination on knowledge distillation's effectiveness in model optimization.

## 2.5 Data Augmentation

Text augmentation is a technique in natural language processing (NLP) that involves artificially expanding a text dataset to improve or augment the data available for training machine learning models. This process is crucial, especially in domains like financial sentiment analysis, where the nuanced interpretation of language can significantly impact model performance. Text augmentation strategies aim to introduce variability and diversity into training datasets, enhancing the model's ability to generalize from limited or skewed data and improving its robustness and accuracy, (Wu Xingand Lv, 2019; Feng et al., 2021; Gong et al., 2022)

In the context of financial sentiment analysis, text augmentation can address challenges such as class imbalance (where positive, negative, and neutral sentiments are not equally represented) and the scarcity of labelled data, which is common in specialized domains. By augmenting the data, researchers can ensure that the model is exposed to a broader spectrum of linguistic expressions and sentiments, leading to improved accuracy in sentiment classification tasks.(Suhaeni and Yong, 2023)

## 2.6 Distillation with Data Augmentation using LLMs

Xu et al., 2024 highlight the critical role of Data Augmentation (DA) through the use of Large Language Models (LLMs) in enhancing Knowledge Distillation (KD) processes. This strategy helps in generating novel, contextually rich training content specifically designed for distinct domains and functionalities. Such an approach capitalizes on the distinctive ability of LLMs to produce data samples that are not only coherent and diverse but also sophisticated, closely mirroring the complex understanding and cognitive capabilities of human experts across various disciplines. Leveraging the foundational knowledge embedded in LLMs, KD utilizes DA to guide LLMs in generating data that encapsulates specialized skills or domain-specific knowledge (West et al., 2022; Chaudhary, 2023). The integration of DA with LLMs significantly enhances the efficiency of the distillation process, enabling the resulting models to develop and refine competencies that would typically demand much larger datasets and considerably greater computational efforts.

In recent studies, (Yoo et al., 2021; Dai et al., 2023) have investigated the application of Large Language Models (LLMs), such as ChatGPT, for the purpose of data augmentation. Combining knowledge distillation with LLM based data augmentation techniques, such as paraphrasing, back translation, and sentence extension, can be effective in improving the performance of distilled models. This strategy effectively mitigates the constraints posed by scarce training data in particular domains or languages, concurrently facilitating improved model generalization.

In summary, while significant advances have been made in the fields of model compression, data augmentation, and domain-specific model training, the integration of these approaches to enhance the performance and efficiency of models like TinyBERT in financial sentiment analysis represents a novel contribution to the field. This research seeks to build upon the foundations laid by previous work, leveraging advancements in knowledge distillation and generative models to address the unique challenges of financial sentiment analysis.

## 2.7 Deployment Challenges with LLM

The deployment of LLMs in real-world applications remains a significant challenge due to their enormous size and computational requirements. For instance, serving a single 175 billion parameter model necessitates at least 350GB of GPU memory, often requiring specialized infrastructure (Zheng et al., 2022). As LLMs grow in size, now surpassing 500 billion parameters, the memory and computational demands escalate further, making them impractical for most product teams, especially those needing low latency performance (Chowdhery et al., 2022).

## 2.8 Summary

This chapter has systematically explored a broad spectrum of scholarly literature foundational to understanding and advancing the fields of sentiment analysis, NLP, and model optimization. Starting with a detailed examination of sentiment analysis, the review highlighted how this area has evolved from simple lexicon-based approaches to sophisticated deep learning models that capture nuanced emotional tones across diverse textual data.

The discussion then transitioned to BERT and its variants, underscoring their important impact on NLP. These models have set new benchmarks in language understanding, leading to a surge in research focused on enhancing their efficiency and applicability through knowledge distillation. This technique, critical for adapting large models to practical use-cases, was thoroughly reviewed, emphasizing methods to retain performance while reducing computational demands.

Data augmentation as a strategy to enrich training datasets and improve model robustness was also covered. The integration of data augmentation with knowledge distillation, particularly using large language models (LLMs), was explored, revealing how these combined approaches enhance model training and performance.

Deployment challenges associated with LLMs, including their size, complexity, and resource requirements, were discussed to address the practical limitations that impact their broader application.

Each section of this review provides a comprehensive overview of current methodologies and their advancements along with a critical assessment of how these techniques can be utilized to overcome practical challenges in deploying advanced NLP systems.

# CHAPTER 3 : RESEARCH METHODOLOGY

## 3.1 Introduction

This chapter delineates the comprehensive methodology employed in this study, detailing the systematic approaches and techniques used to conduct the research. The chapter is structured to first introduce the research design, which outlines the overall strategy and logical sequence for integrating and analysing the datasets involved in this study. Following this, the chapter describes the datasets utilized, including the Financial PhraseBank, FiQA 2018 Task1, and Forex News Annotated datasets, each crucial for evaluating the performance of the language models discussed.

Subsequent sections delve into the synthetic data generation strategies employed to augment these datasets using advanced large language models (LLMs). This includes generating new training examples, labelling these examples, and creating variations of mislabelled sentences to enhance model training efficacy. Additionally, the approach for generating new unlabelled examples is presented, highlighting the cost-effective use of LLMs to expand the training corpus without the need for explicit labelling.

Model development and fine-tuning procedures are comprehensively covered, beginning with an overview of Transformers, followed by a detailed examination of BERT and its financial adaptation, FinBERT. This study introduces Augmented FinBERT, an enhanced version of FinBERT developed using synthetic data, and describes the fine-tuning strategies that ensure its robustness and effectiveness in financial sentiment analysis.

Performance metrics and their rationale are outlined to evaluate the effectiveness of the models developed. The methodology for measuring these performance metrics is also specified, ensuring a clear understanding of how model outcomes are assessed.

Finally, the knowledge distillation strategies employed to transfer sophisticated capabilities from Augmented FinBERT to TinyFinBERT are discussed. This includes a detailed explanation of the steps involved in distilling knowledge effectively, ensuring that TinyFinBERT inherits the nuanced understanding necessary for accurate sentiment analysis despite its smaller size.

Through this methodological framework, the study aims to demonstrate the effectiveness of LLM augmented data in enhancing the accuracy and efficiency of financial sentiment analysis models.

## 3.2 Research Design

This research employs a quantitative approach, integrating both experimental and computational methods to address the complex challenges of financial sentiment analysis. The study is structured into distinct phases: data preparation, model training and optimization, knowledge distillation, and model evaluation. Each phase utilizes specific computational techniques to explore and analyse the impact of synthetic data and advanced modeling strategies on the performance of financial sentiment analysis models.

1. **Experimental Design:** The core of the experimental approach involves generating synthetic training data using state-of-the-art generative models, GPT-4 Omni and GPT-3.5 Turbo. This process tests the hypothesis that synthetic data can effectively augment existing datasets, thereby enhancing the training of FinBERT. Subsequently, the augmented FinBERT model undergoes a series of experiments to optimize its parameters and settings to best utilize the enhanced dataset.

2. **Computational Methods:** The computational aspect of the study includes the application of sophisticated machine learning algorithms and knowledge distillation techniques. This involves the use of automated scripts and frameworks to fine-tune and distill models, ensuring that each iteration is consistent and replicable. The use of computational methods extends to the evaluation phase, where models are tested against unseen data to assess their generalization capabilities.

**Justification for Research Design**

The selection of a quantitative, experimental, and computational research design is aligned with the challenges and objectives outlined in the study's aim. The integration of these methods addresses several key challenges:

**Data Scarcity:** One of the primary challenges in training specialized models like FinBERT is the limited availability of high-quality, domain-specific training data. The experimental creation and incorporation of synthetic data generated by LLMs provide a novel solution to this problem, enabling the enhancement of the model's learning and generalization capabilities.

**Computational Efficiency:** By utilizing computational methods to implement knowledge distillation strategies, this research design directly contributes to reducing the computational demands of deploying high-performance models. The design allows for the systematic exploration of different distillation parameters and their effects on model efficiency and performance.

**Model Accuracy and Generalization:** The quantitative nature of the study ensures that results are measurable, and repeatable. This approach allows for rigorous testing of the hypothesis that knowledge distillation, supported by synthetic data, can maintain or even enhance the accuracy of smaller, more efficient models like TinyFinBERT in the financial domain. Further, the use of unseen datasets in the evaluation phase tests the models' ability to generalize, thus providing a robust assessment of their practical utility in real-world settings.

**Real Time Analysis Capability in Resource Constrained Environments:** The overall design is aimed at developing models that not only perform well on standard benchmarks but are also viable in environments where computational resources are limited. This is crucial for deploying AI solutions in real-world financial markets, where speed and efficiency are as important as accuracy.

By planning and justifying each element of the research design, this study ensures that all objectives are addressed, leveraging both experimental and computational methods in order to provide valuable insights relevant to the field of financial sentiment analysis.

## 3.3 Dataset Description

### 3.3.1 Financial PhraseBank Dataset

The Financial PhraseBank dataset[1], developed by (Malo et al., 2014), is a comprehensive collection of 4,846 sentences sourced from financial news texts and company press releases, found through the LexisNexis database. This dataset serves as an important resource for training and testing in the domain of financial sentiment analysis.

**Composition and Annotation**

Each sentence in the PhraseBank has been evaluated and categorized into positive, negative, or neutral categories, reflecting the sentiment conveyed regarding the financial outlook of companies mentioned. This categorization was performed by a panel of 16 annotators who have adequate background knowledge on financial markets (Malo et al., 2014). The annotator agreement levels were also recorded. This results in multiple subsets of the data that represent different levels of consensus. The PhraseBank dataset comprises of sentences with at least 50% agreement levels.

For this study, we employ all the records in the PhraseBank which corresponds to sentences with at least 50% agreement. This is consistent with the approach used for FinBERT. This ensures a fair comparison and additionally a merged dataset that combines all levels of agreement, offers a broader spectrum of data variability and robustness. The file "Sentences_50Agree.txt" corresponding to at least 50% agreement in the PhraseBank dataset contains 4846 records. These categorizations are detailed in Section 4.2.1 below.

**Relevance to the Study**

The Financial PhraseBank dataset has been instrumental in the development and fine-tuning of the FinBERT model, as described by (Araci, 2019). Its extensive use in academic research for financial sentiment analysis underscores its utility and credibility. Some of the prior works where this dataset was used include (Yang et al., 2020; Liu et al., 2024; Peng et al., 2024). In our study, this dataset along with LLM augmented data is used in the training process for Augmented FinBERT and TinyFinBERT. By providing a well-annotated, domain-specific corpus, the PhraseBank dataset enables effective training and rigorous evaluation under controlled conditions, mirroring the methodology used in the original FinBERT study.

---

[1] https://www.researchgate.net/publication/251231364_FinancialPhraseBank-v10

**Utilization in Current Research**

Consistent with the original FinBERT study, we partition the dataset by setting aside 20% of the sentences to form a test set. The test set comprises of 970 records. The remaining 3876 records are used for training and data augmentation. The final model performance metrics reported in this study are calculated exclusively on the test data. This test data is used solely for final evaluation and has not been utilized in any phase of training the FinBERT model, ensuring an independent and unbiased assessment.

### 3.3.2  FiQA 2018 Task1 Dataset

The FiQA 2018 Task1 (Maia et al., 2018) dataset[2] was developed for the WWW'18 conference focused on financial opinion mining. It comprises annotated financial data sourced from platforms like StockTwits, Reddit, Wikinews, and various other financial websites. This dataset includes a total of 438 financial headlines and 675 financial tweets in the training set, making it a rich resource for assessing sentiment analysis models. These 1113 records are used in this study.

**Composition and Annotation**

In the FiQA dataset, each tweet or headline is associated with a specific financial entity and is annotated with a continuous sentiment score ranging from -1 (very negative) to +1 (very positive). This granularity allows for a nuanced understanding of sentiment in the context of financial discussions. For the purposes of this study, these continuous sentiment scores have been converted into three discrete categories to align with the Financial PhraseBank's format as described in Section 4.2.2.

**Relevance to the Study**

The FiQA 2018 Task1 dataset's close similarity to the Financial PhraseBank in terms of content but differing in sentiment scoring (continuous vs. discrete) provides a unique opportunity to validate the robustness and adaptability of the sentiment analysis models developed in this research. It serves as an excellent benchmark for measuring the generalization capabilities of both the Augmented FinBERT and TinyFinBERT models, particularly in how well these models can adapt their learning from one sentiment annotation style to another and perform across various financial platforms and formats. Similar to the Financial PhraseBank dataset, this dataset is also used extensively in academic research for financial sentiment analysis. Some of the prior works where this dataset was used include (Yang et al., 2020; Shang et al., 2023; Liu et al., 2024). This dataset was also used in the original FinBERT (Araci, 2019) paper.

---

[2] https://sites.google.com/view/fiqa

**Utilization in Current Research**

This dataset is employed to test the generalization capabilities of the models trained using the Financial PhraseBank dataset. By assessing model performance on an independent but conceptually similar dataset, we can better understand the models' strengths and limitations in real-world applications. The FiQA dataset, with its mixed format of tweets and headlines and continuous sentiment scores, challenges the models to adapt their learning to different types of financial discourse and more granular sentiment distinctions. This use not only highlights the flexibility and scalability of the models but also tests their effectiveness in a broader context of financial sentiment analysis.

### 3.3.3   Forex News Annotated Dataset

The Forex News Annotated dataset, created by (Fatouros et al., 2023), consists of 2,291 financial news headlines from the year 2023, specifically focusing on major forex pairs such as AUDUSD, EURCHF, EURUSD, GBPUSD, and USDJPY. These headlines were sourced from well-established financial news platforms, Forex Live and FXstreet, over a span of 86 days between January and May. This dataset is distinct because it targets the forex market, which can exhibit sentiment dynamics and reactions different from other sectors of the financial market.

**Composition and Annotation**

Each headline has been annotated with sentiments based on the expected short-term impact on the forex pair value, categorizing into 'positive', 'negative', and 'neutral'. These sentiments correspond to bullish, bearish, and hold market sentiments respectively, offering insights into the forex market's specific reactions to news:

**Positive:** Indicative of bullish market sentiments, suggesting an anticipated increase in the value of the forex pair.

**Negative:** Reflective of bearish market sentiments, suggesting a likely decline in the value of the forex pair.

**Neutral:** Corresponding to stable or hold market sentiments, where no significant immediate impact on the forex pair value is anticipated.

This three-level labelling system mirrors the sentiment classification used in the Financial PhraseBank, allowing for consistent sentiment analysis across diverse datasets.

**Relevance to the Study**

The Forex News Annotated dataset provides a unique opportunity to evaluate the models' generalization capabilities beyond the traditional financial news context, such as that found in the Financial PhraseBank and FiQA datasets. Unlike these datasets, the Forex News Annotated dataset focuses exclusively on the forex market, which might not correlate as directly with the broader financial data the FinBERT model was initially trained on. This differentiation is critical as it challenges the models to adapt to and accurately analyse sentiments within a specialized and potentially divergent financial domain.

**Utilization in Current Research**

In this study, the Forex News Annotated dataset is utilized to test the adaptability and effectiveness of both Augmented FinBERT and TinyFinBERT in new and specific market conditions not previously encountered during their training. By incorporating this dataset, the research aims to rigorously assess how well these models can generalize their learning to different and potentially non-correlated financial data, purely to evaluate their robustness and flexibility.

## 3.4 Synthetic Data Generation Strategy

To address challenges associated with sourcing labelled data in specialized domains and to transfer advanced language understanding capabilities from GPT-4 Omni to FinBERT, this study employs synthetic data generation strategies using LLM. The aim is to enhance the training of the FinBERT model, making it more robust and versatile. This is achieved by generating new synthetic examples specifically targeting areas where FinBERT initially struggled, thereby improving both accuracy and generalizability.

**Strategy for Generating Synthetic Data**

(He et al., 2023) in their research identified that even high-performing NLP models can exhibit systematic failures on specific subgroups within data. These subgroups often go unnoticed because they are underrepresented in the training data available, and traditional methods of data collection do not necessarily rectify this imbalance. Their framework uses LLMs with human oversight to identify and generate data for these challenging subgroups, significantly improving model accuracy and fairness. Similar to this approach, our study uses GPT-4 Omni to generate synthetic data targeting mislabelled instances by FinBERT. This targeted synthetic generation directly addresses the known weaknesses of FinBERT, enhancing both the robustness and accuracy of its sentiment analysis.

**Identification of Model Weaknesses**

The need to enhance FinBERT's performance on subtle sentiment distinctions emerges from observed inconsistencies in its ability to accurately differentiate between positive and neutral statements. (Malo et al., 2014) noted significant annotator disagreements in these classifications, with the agreement rates for positive-neutral distinctions notably lower at 75.2% compared to other sentiment pairings. Such discrepancies often stem from the challenges in distinguishing generic positive expressions, often referred to as "company glitter", from genuinely positive sentiments regarding a company's performance.

Supporting this observation, the confusion matrix provided by (Araci, 2019) (see Figure 3.1) presents clear evidence of specific challenges faced by FinBERT. It shows that 6% of the instances where neutral statements were incorrectly classified as positive and 4% where positive statements were mislabelled as neutral. These misclassifications underscore FinBERT's difficulties with the nuanced boundaries between neutral and positive sentiments.

|  | Predicted | | |
|---|---|---|---|
| Actual value | Positive | Negative | Neutral |
| Positive | 0.2 | 0.01 | 0.04 |
| Negative | 0.01 | 0.11 | 0.01 |
| Neutral | 0.06 | 0.02 | 0.55 |

**Figure 3.1: Confusion Matrix for original FinBERT model**

Note: Due to rounding off, numbers do not sum to 1. Source: (Araci, 2019).

Additionally, it can be seen that for 2% of the instances, neutral statements were incorrectly classified as negative. To address these issues, we employ a targeted prompting strategy with GPT-4 Omni to generate new examples that specifically challenge the model's weak areas. This approach focuses on generating boundary cases that FinBERT has historically mislabeled, thereby directly improving its ability to discern between closely tied sentiment categories.

### 3.4.1 Using LLM to Generate New Training Examples

To address the identified weaknesses in FinBERT model, this study employs GPT-4 Omni to generate synthetic examples that target these challenging boundary areas. The use of a sophisticated LLM like GPT-4 Omni allows for the creation of high-quality, nuanced examples in areas that FinBERT previously misclassified. This approach aligns with the methodology described in recent academic contributions such as (He et al., 2023), where Targeted Data Generation (TDG) is used to enhance model performance on specific challenging subgroups. By simulating complex scenarios that FinBERT is likely to encounter in real-world applications, these synthetic examples are tailored to improve the model's accuracy and reliability in distinguishing closely related sentiment categories.

**Prompt Engineering for Synthetic Generation**

The implementation involves strategically prompting GPT-4 Omni to produce financial statements that hover on the fine line between positive and neutral sentiments. These prompts are designed to elicit responses that embody the ambiguity found in real financial communications, thus providing FinBERT with the opportunity to learn from and adapt to these subtleties. The process not only enhances FinBERT's sensitivity to slight variances in sentiment but also ensures that it becomes more adept at handling similar challenges autonomously in the future. In our study, the OpenAI provided API was utilized to generate new synthetic examples. The settings chosen for the API calls were carefully tailored to ensure the generation of high-quality, relevant financial sentences:

- **Temperature (0.6):** The temperature setting of 0.6 was selected to strike a balance between creativity and relevance. A lower temperature might restrict the model to more predictable outputs, while a higher setting could lead to less reliable and more diverse outputs. At 0.6, the model produces sentences that are innovative yet closely aligned with the typical patterns observed in financial news language.
- **Top_p (1):** Setting top_p to 1 ensures that the model considers the full range of possible next words while generating text. This setting is chosen to maximize the potential diversity of the output without constraining the word selection process.

- **Frequency Penalty (0.2):** This setting reduces the likelihood of repeating the same information, which is crucial when generating multiple examples where varied expressions are needed to capture nuanced sentiment differences without redundancy.
- **Presence Penalty (0):** A zero-presence penalty means the model does not avoid using words it has already used, which is important in financial contexts where specific terminology may need to be repeated across different sentences for clarity and accuracy.

**Prompt Design**

To generate synthetic data that enhances the FinBERT model's accuracy and generalizability, specific prompts were meticulously crafted. These prompts are designed to elicit financial news sentences that closely mimic the subtle nuances and complexities encountered in real-world financial reporting. The main system prompt used encapsulates the depth and precision required for this task:

**System Prompt:**

*"As a financial expert with an in-depth understanding of stock market dynamics, you're adept at analyzing how news headlines can subtly influence company stock prices. Your task is to generate financial news sentences that accurately maintain their original sentiment—positive, negative, or neutral. Each sentence should subtly reflect potential impacts on stock prices, preserve contextual integrity, and maintain factual accuracy. The language should be sophisticated, suitable for financial reporting, and resonate with professional investors, market analysts, and financial reporters."*

This prompt sets the stage for generating sentences that are not only factually accurate and contextually intact but also nuanced in sentiment, reflecting the complexity of real financial discourse.

**Examples of User Prompts**

To further refine the data generation process, several user-specific prompts were employed to target different sentiment classifications, particularly focusing on the boundary cases where FinBERT has shown weaknesses. Here are examples of such prompts:

**Neutral Sentiment with Subtle Implications:**

**Prompt:**

*"Generate 50 different financial statements regarding a company's quarterly earnings with neutral sentiment. Each statement should subtly imply future prospects without definitive positive or negative forecasts. Use financial terminology that conveys uncertainty or cautious optimism. Each statement must be concise, under 50 words, and avoid overtly optimistic or pessimistic language, focusing instead on providing a balanced view."*

**Example Output:**

"1. The company's quarterly earnings met expectations, suggesting steady performance amidst market fluctuations.

…"

**Neutral Potentially Viewed as Negative:**

**Prompt:**

*"Develop 50 unique financial news statements about a company's quarterly performance, each formulated to be neutral but potentially viewed negatively at first glance. Refrain from using any definitive positive or negative terms, focusing on neutral, unbiased descriptions."*

**Example Output:**

"1. This quarter, the company experienced slight disruptions in supply chain operations, which were managed without major financial repercussions.

…"

**Positive Potentially Viewed as Neutral or Negative:**

**Prompt:**

"Develop 50 financial communications regarding a company's quarterly financial performance. Each statement should be formulated to be positive but potentially viewed as neutral or negative at first glance. Avoid explicit positive words, keep each statement concise, integrating critical financial analysis terms."

**Example Output:**

"1. There was a significant reduction in operational downtimes this quarter, potentially leading to higher productivity and earnings.
…"

Through trial and error multiple variations of such prompts were used to generate new sentences. These prompts are integral to the methodology as they directly address the nuanced challenges identified in FinBERT's performance. By focusing on generating synthetic examples around these specific boundary conditions, the study not only aims to improve the model's accuracy on challenging subgroups but also enhances its ability to generalize across different financial contexts.

### 3.4.2  LLM Labelling of Generated Examples

After generating the synthetic examples, GPT-4o was also employed to label these sentences with sentiments, using a temperature of 0 to ensure deterministic, consistent outputs. The utilization of GPT-4o for sentiment labelling in this study is supported by findings from (Belal et al., 2023), who demonstrated that ChatGPT, a precursor to GPT-4o, significantly outperforms lexicon-based unsupervised methods in labelling accuracy across various sentiment analysis tasks. This research underscores the advanced capabilities of GPT models in processing and understanding complex textual data, suggesting that GPT-4o is likely to offer similar, if not enhanced, efficacy in accurately classifying sentiments in financial texts.

Leveraging the insights from (Belal et al., 2023), this thesis employs GPT-4o to perform labelling of synthetic financial statements. The proven effectiveness of GPT models in sentiment classification justifies their use in enhancing the data quality for FinBERT training, aiming to improve both the model's robustness and its analytical precision in financial sentiment analysis.

**System Prompt for Sentiment Labeling**

The labelling was performed in batches of 100 examples per API call to comply with operational constraints and manage the API usage efficiently.

To ensure that the sentiment labelling process mirrored the analytical depth required for financial sentiment analysis, a comprehensive system prompt was devised. This prompt reflects the need for a nuanced understanding of market dynamics and the potential impact of news on stock prices. The prompt used is as follows:

**System Prompt:**

*"As a seasoned financial analyst well-versed in the intricacies of market dynamics, you are adept at discerning the subtle nuances in news statements that can sway investor sentiment and assess the impact of such news statements on stock prices. Your task is to analyse the potential impact of financial news statements on the stock prices of the relevant companies. Assess whether each statement suggests a positive, negative, or neutral effect on the company's stock price. Ensure that your analysis of each statement is conducted independently to avoid bias from other statements."*

Each batch of synthetic examples was processed using a user-specific prompt that focused on the individual analysis of each statement, emphasizing independent evaluation to avoid contextual bias. This prompt was structured to ensure that each analysis was both thorough and isolated, considering only the information presented within each specific statement. The user prompt is as follows:

**User Prompt:**

*"As a seasoned financial analyst, you are to analyse the following 100 numbered independent financial news statements, each pertaining to a different company. Assess the potential impact on the stock price of the relevant company independently of any other statements, even when they mention the same company name. Determine whether each statement suggests a positive, negative, or neutral effect on the company's stock price. Consider factors such as tone, relevance of information, and market context in your evaluation. Please only provide the potential impact (positive, negative, or neutral) for each statement in a numbered format, with each prediction corresponding to the order of the statements provided."*

The rationale behind employing these detailed prompts is to replicate the decision-making process a financial analyst might go through when interpreting news for investment insights. By instructing GPT-4o to adopt the role of a financial analyst, the model is better positioned to apply its advanced natural language understanding capabilities to accurately predict the sentiment implications for stock prices. This approach not only aligns with the synthetic data generation goals but also enhances the reliability of the sentiment labels assigned to each statement, ensuring that the labels are reflective of thoughtful financial analysis rather than superficial interpretation.

To ensure the highest level of accuracy and reliability in the sentiment labels assigned by GPT-4o, the following strategies were implemented:

**Multiple Iterations**

All the synthetic examples were processed five times, with the order of the data randomly shuffled in each iteration. This measure was designed to mitigate any potential bias or sequence effects that could affect consistency in labelling. The iterative process ensures that each label is robust and that the sentiment classification is not influenced by the positioning or grouping of sentences within a batch.

**Selection Criteria for Training**

For an example to be included in the final training set for FinBERT, it must receive the same label in all five iterations, a rigorous criterion that guarantees only the most consistent and reliable examples influence the model's learning. This stringent selection process is crucial for refining FinBERT's ability to accurately interpret and analyse financial sentiments, particularly in ambiguous or boundary cases. Utilizing GPT-4 Omni to automatically annotate these generated sentences with sentiment labels, ensures consistency and accuracy through controlled settings and multiple iterations.

**Use of Mismatched Examples**

Examples that received inconsistent labels across different iterations were not discarded. Instead, they were utilized as unlabelled data in the knowledge distillation process for TinyFinBERT. This approach allows TinyFinBERT to benefit from complex and varied data, enhancing its ability to generalize from ambiguous cases and improve its analytical depth. The inclusion of such data ensures that TinyFinBERT's training environment is richly diverse, potentially increasing its adaptability and effectiveness in real-world applications.

The high-quality, challenging examples generated are used to train Augmented FinBERT, focusing specifically on improving its ability to accurately parse and interpret complex financial texts. This targeted training helps in fine-tuning FinBERT's sensitivity to subtle nuances in financial discourse, greatly enhancing its analytical precision and generalization capabilities across diverse financial scenarios.

Through this strategic use of LLM augmentation, the study harnesses the power of advanced generative models to significantly boost the performance of traditional sentiment analysis models like FinBERT. By integrating LLM-generated examples into FinBERT's training regimen, we aim to bridge the gap between state-of-the-art language models and specialized financial sentiment analysis tools, thereby advancing the field and providing more efficient tools for financial market analysis.

### 3.4.3 Using LLM to Create Variations of Mislabeled sentences

The utilization of GPT-4o for generating variations of existing sentences is supported by findings from Dai et al. (2023) and Giovanni Møller et al. (2024), who have shown the successful use of LLMs like ChatGPT for similar data augmentation.

Similar to the process used for generating new examples, the process of creating variations also employs a targeted approach aimed at improving FinBERT's handling of data where initial sentiment classification proved challenging. This process involves identifying sentences that were incorrectly labeled by FinBERT and subsequently verified with GPT-4o to ensure that the generated variations align with the correct sentiment labels recognized by both human annotators and the LLM.

**Systematic Identification and Labeling**

Initially, sentences mislabelled by FinBERT are identified, then relabeled by GPT-4o using a stringent five-iteration process, similar to the methodology described for initial labelling. Only sentences where GPT-4o's labelling matches the original dataset labels are retained. This ensures that the selected sentences for variation generation are those where there is a consensus on sentiment, thereby maintaining the integrity of the training data.

**Prompt Design for Variation Generation**

To generate the sentence variations, the following system and user prompts were designed:

**System Prompt:**

*"As a seasoned financial analyst well-versed in the intricacies of market dynamics, you are adept at discerning the subtle nuances in news statements that can sway investor sentiment and assess the impact of such news statements on stock prices. Your task is to generate 10 variations of an existing financial news statement that effectively convey the original sentiment — positive, negative, or neutral — reflecting the potential impact on stock prices. Each variation must maintain the sentiment of the original sentence but use different words or structures. Employ diverse language and sentence structures, and integrate domain-specific terminology to ensure authenticity and relevance."*

**Rationale:**

- **Role-playing as a financial analyst:** This setup leverages the language model's ability to adapt to specific user personas, which in this case, is a financial analyst. This encourages the model to generate content that aligns with professional financial analysis, ensuring that the language and sentiment are appropriate for the context.
- **Task specificity:** The prompt clearly defines the task, generating variations that maintain the original sentiment. This specificity helps guide the language model to focus on sentiment preservation while allowing for linguistic diversity, crucial for training data augmentation.
- **Integration of domain-specific terminology:** By explicitly requesting the use of specialized financial language, the prompt ensures that the generated sentences are not only diverse but also contextually relevant and authentic to the field of finance.

**User Prompt:**

*"In your role as a seasoned financial analyst, generate 10 variations of the following financial news statement "{row['text']}" while retaining the "{row['label']}" sentiment and financial context. Incorporate diverse language and sentence structures, and integrate domain-specific*

*terminology to ensure authenticity and relevance within the financial sector. Aim for variations of similar length and complexity to the original sentence. Please avoid using unreadable characters. Provide only the 10 numbered statements without explanations."*

**Rationale:**

- **Intentional Repetition of items in System Prompt:** The intentional repetition of items in the user prompt that have already been mentioned in the system prompt reinforces the critical requirements and focuses the language model's output, ensuring consistency and clarity in executing the specific task of generating high-quality, contextually appropriate variations.
- **Preservation of sentiment and financial context:** Mentioning the original sentiment alongside the sentence for which variations are to be generated ensures that the language model maintains the intended emotional tone and context in each variation, crucial for accurately training sentiment analysis models on consistent and reliable data.
- **Operational specifics:** Specifying the number of variations, the avoidance of unreadable characters, and the format for responses ensures that the output is structured and usable immediately for training purposes with limited preprocessing required.

**Example of Application**

Original Sentence with Neutral Label:

*"According to statistics by the Finnish Maritime Administration, shipping did not increase much in January-August 2009, however."*

10 Generated Variations:

1. *As per data from the Finnish Maritime Administration, there was minimal growth in shipping activities during January to August 2009.*
2. *The Finnish Maritime Administration's statistics indicate that shipping volumes saw little change from January through August 2009.*
3. *Figures released by the Finnish Maritime Administration show that shipping remained relatively stable in the January-August 2009 period.*

4. *Based on the Finnish Maritime Administration's reports, there was only a slight uptick in shipping between January and August 2009.*
5. *The Finnish Maritime Administration data reveals that there was no significant increase in shipping from January to August 2009.*
6. *According to the Finnish Maritime Administration's statistics, the shipping sector experienced only marginal growth during January-August 2009.*
7. *Statistics provided by the Finnish Maritime Administration suggest that shipping levels hardly increased from January through August 2009.*
8. *The Finnish Maritime Administration's figures for January to August 2009 show that there was not much of an increase in shipping activity.*
9. *Data from the Finnish Maritime Administration indicates that there was negligible growth in shipping activities over the period of January to August 2009.*
10. *According to reports by the Finnish Maritime Administration, there was only a modest rise in shipping between January and August of 2009.*

**Validation and Use of Variations**

The variations generated are subsequently relabeled by GPT-4o, undergoing another five iterations to ensure consistency and reliability in sentiment classification. This validation step is crucial to guarantee that only the most accurate variations are used to further finetune FinBERT.

**API Configuration Rationale for Synthetic Sentence Generation**

The API configuration settings for generating synthetic sentence variations using a language model like GPT-4o are crucial for achieving appropriate variations that are contextually relevant. The explanation of specific parameter used and the rationale behind its selection are provided below:

**Temperature (0.7):**

**Purpose:** Controls the randomness of the response generated by the language model.

**Rationale:** A temperature setting of 0.7 strikes a balance between creativity and control in the model's responses. It allows for a moderate level of variation in word choice and sentence structure without deviating too far from the logical and contextual framework of the original sentence. This level ensures that the variations are neither too predictable nor too divergent, which is essential for maintaining the original sentiment while introducing necessary diversity.

**Max Tokens:**

**Purpose:** Defines the maximum length of the generated text.

**Rationale:** Setting this parameter according to the length of the original sentence ensures that the variations maintain a similar complexity and information density. This consistency is important for training purposes, as it allows the model to learn from examples that reflect the typical length and detail of financial statements without introducing an unintended bias toward shorter or longer texts. Please note that the prompt also specifically asks the LLM to generate sentences of similar length. This setting acts as a second check.

**Top_p (1):**

**Purpose:** Determines how the model samples from its vocabulary based on the cumulative probability distribution.

**Rationale:** A top_p of 1 effectively disables the nucleus sampling, meaning the model considers the full range of vocabulary when generating each token. This setting is chosen to maximize the potential diversity of the output without artificially restricting the model's vocabulary choices, which is vital for capturing the full spectrum of financial terminology and expression styles.

**Frequency Penalty (0.5):**

**Purpose:** Decreases the likelihood of repeatedly using the same word in the generation.

**Rationale:** Applying a moderate frequency penalty helps to ensure that the generated variations are not redundant and that they use a wider range of vocabulary. This is particularly important in financial reporting contexts, where nuanced differences in terminology can significantly alter the perceived sentiment of a statement. The penalty helps diversify the linguistic structure without losing the original message's intent.

**Presence Penalty (0):**

**Purpose:** Adjusts the model's tendency to avoid repeating topics or themes already mentioned in the conversation.

**Rationale:** Setting the presence penalty to zero means that the model does not avoid reusing words that have appeared in the interaction. This is crucial when the accuracy and consistency of financial terminology are needed across generated variations. It ensures that key terms necessary for maintaining the statement's factual integrity are not omitted.

The use of a moderate frequency penalty (0.5) in conjunction with a zero presence penalty in the study ensures that while the LLM avoids unnecessary repetition of the same words, and doing so enhances the lexical diversity of the generated text, it does not overly avoid reiterating crucial financial terms. This combination helps in generating synthetic financial texts where each variation is unique and yet maintains essential terminologies.

The chosen API configuration is designed to optimize the balance between creative diversity and contextual fidelity in the generated text. These settings ensure that the synthetic sentences produced are suitable for training FinBERT, offering varied yet accurate reflections of potential real-world financial statements. This approach enhances the model's ability to generalize and learn from high quality examples generated by the LLM.

### 3.4.4  Using LLM to Generate New Unlabeled Examples

To efficiently generate a large volume of unlabelled data necessary for knowledge distillation, we utilize GPT-3.5 Turbo API in batch mode. This choice is primarily driven by cost-effectiveness, as GPT-3.5 Turbo (gpt-3.5-turbo-0125) is 10 times cheaper than GPT-4 Omni (gpt-4o-2024-05-13), making it a practical option for generating extensive augmented data without the immediate need for precise labelling or sentiment consistency.

**Data Generation Strategy:** The strategy for creating new examples follows the method detailed in Section 3.4.1, with the principal modification being the use of GPT-3.5 Turbo instead of GPT-4 Omni. This ensures a consistent approach in data generation while optimizing resource allocation. For generating variations, we employ the same technique as described in Section 3.4.3. However, instead of limiting variations to mislabelled data, variations are now produced for all examples in the training dataset. Despite this broader application, a greater number of variations are still specifically generated for mislabelled data to better support the training process during knowledge distillation. The initial new sentences generated as per Section 3.4.1 serve as seed knowledge for the LLM, prompting it to create further variations.

**Sentiment Consistency:** While the LLM prompts are designed to maintain the original sentiment during the generation of new examples, this specific aspect of sentiment consistency is not rigorously verified post-generation. This deliberate methodology is aligned with the primary objective of the data augmentation phase, which focuses on significantly increasing the volume of training data available for the knowledge distillation process. At this stage, the emphasis is on exposing the student model to a broader range of linguistic contexts and variations, rather than on the precise calibration of sentiment accuracy. This strategy is beneficial as it allows the student model to encounter and adapt to a wide spectrum of data scenarios, thus fostering a more robust generalization capability rather than restricting learning to highly curated sentiment accuracy, which can be refined in subsequent targeted training phases.

Furthermore, as outlined in Section 3.4.2, examples that received inconsistent labels and were not employed as labeled data have been utilized as unlabelled data in this phase of the knowledge distillation process for TinyFinBERT. Employing these examples as unlabelled data not only enriches TinyFinBERT's training environment with a diverse array of inputs, enhancing its adaptability and effectiveness in real-world scenarios, but also proves to be cost-effective. This approach maximizes the use of existing data resources, reducing the need for additional data generation and processing.

### 3.5 Model Development and Finetuning

#### 3.5.1 Transformers

Transformers are a class of deep learning models pivotal to modern natural language processing. Introduced in the paper "Attention is All You Need" by (Vaswani et al., 2017), Transformers eschew traditional sequential mechanisms in favor of the self-attention mechanism that processes all words in a sentence concurrently, allowing for more substantial modeling of dependencies without regard to their distance in the input sequence.

Most recent pre-trained language models such as BERT, RoBERTa, T5, and the Generative Pre-trained Transformer series developed by OpenAI, including GPT-2, GPT-3, and the latest GPT-4, employ Transformer layers. These layers are adept at capturing long-term dependencies between input tokens via the self-attention mechanism. A standard Transformer layer consists of two primary sub-layers: multi-head attention (MHA) and a position-wise feed-forward network (FFN).

**Multi-Head Attention (MHA):**

The attention mechanism is structured around three key components: queries, keys, and values, denoted as matrices Q, K, and V. The attention scores are computed as

$$\boldsymbol{A} = \frac{\boldsymbol{Q}\boldsymbol{K}^T}{\sqrt{d_k}} \qquad (1)$$

$$\text{Attention}(\boldsymbol{Q}, \boldsymbol{K}, \boldsymbol{V}) = \text{softmax}(\boldsymbol{A})\boldsymbol{V} \qquad (2)$$

where $\sqrt{d_k}$ normalizes the scores to prevent instability in the softmax step that follows, which is applied across the rows of $\boldsymbol{A}$. This normalization is crucial for maintaining stable gradients during training. The output is the weighted sum of values, where the weights are the softmax-normalized scores, enhancing the model's ability to focus on relevant parts of the input sequence.

According to (Clark et al., 2019; Jiao et al., 2019), BERT's attention matrices can encapsulate significant linguistic knowledge, essential for our distillation strategy. Multi-head attention improves the model's capacity by projecting $\boldsymbol{Q}, \boldsymbol{K},$ and $\boldsymbol{V}$ to different representational subspaces across multiple heads:

$$\text{MHA}(\boldsymbol{Q}, \boldsymbol{K}, \boldsymbol{V}) = \text{Concat}(h_1, \ldots, h_k)\boldsymbol{W} \qquad (3)$$

Here, $k$ denotes the number of attention heads, each contributing to a broader understanding of the input.

**Position-wise Feed-Forward Network (FFN):**

Each Transformer layer further contains an FFN, which consists of two linear transformations with a ReLU activation between them:

$$\text{FFN}(x) = \max(0, x\boldsymbol{W}_1 + b_1)\boldsymbol{W}_2 + b_2 \qquad (4)$$

The ReLU activation introduces non-linearity, enabling the network to compute more complex functions and adding depth to the model's capabilities.

**Key Features of Transformers:**
- **Self-Attention Mechanism:** This core feature allows each word to dynamically adjust its influence based on other words in a sentence, greatly enhancing the model's ability to understand context.
- **Parallel Processing:** Unlike their predecessors (RNNs and LSTMs), Transformers process all words simultaneously, which drastically reduces training times and enhances model scalability.
- **Scalability and Efficiency:** Due to their ability to perform computations in parallel, Transformers are particularly well-suited for training over large datasets, a common scenario in tasks requiring extensive knowledge like those in NLP.
- **Adaptability:** The architecture's flexibility enables extensive pre-training on general tasks, followed by fine-tuning on specific datasets, a method that has proven effective across various NLP applications. Models like BERT and GPT are examples of Transformers that have been pre-trained on a large corpus.

### 3.5.2 BERT

BERT (Bidirectional Encoder Representations from Transformers) (Devlin et al., 2019), a groundbreaking model built on the Transformer architecture, redefines the approach to pre-training language models. Unlike traditional models that predict the next word in a sequence, BERT learns to predict words that have been intentionally masked in a sentence, making it inherently bidirectional. This technique, known as Masked Language Modeling (MLM), allows the model to freely learn the context of a word based on all its surroundings (left and right context), which is not possible in unidirectional models.

Some features of BERT are as follows:
- **Masked Language Modeling (MLM):** During training, BERT masks 15% of the tokens in each sequence at random and then predicts these masked tokens. This approach allows the model to develop a deep, bidirectional understanding of language context.
- **Next Sentence Prediction (NSP):** BERT is also trained to predict whether two given sentences follow one another in a document. This task enables the model to capture relationships between consecutive sentences and enhances its ability to handle tasks requiring an understanding of the text structure.
- **Input Representation:** BERT combines token and position embeddings with special tokens ([CLS] for classification tasks and [SEP] to separate segments) to handle a variety of tasks from simple classification to question answering.

### 3.5.3 FinBERT

BERT's ability to model language context deeply and comprehensively is crucial for sentiment analysis, particularly in domains like finance where the context can significantly influence the sentiment expressed. FinBERT (Araci, 2019), a variant of BERT fine-tuned specifically for financial text, leverages this architecture to better understand and interpret complex financial narratives.

**Pre-training on Financial Corpus**

To adapt its capabilities to the financial sector, FinBERT is pre-trained on a substantial financial corpus known as TRC2-financial, a subset of Reuters' TRC24. This dataset includes around 1.8 million news articles published between 2008 and 2010. The pre-training on this specific corpus allows FinBERT to internalize the lexicon, syntax, and thematic content typical of financial texts, thus making it adept at handling the complexities and subtleties of financial language.

**Fine-tuning for Sentiment Analysis**

After pre-training, FinBERT undergoes a fine-tuning process on the Financial PhraseBank dataset, which is specifically labeled for sentiment analysis. This dataset enables FinBERT to learn the various expressions of sentiments within financial contexts, improving its accuracy in classifying sentiments in new financial texts. The fine-tuning involves appending a dense layer to process the output from the [CLS] token, traditionally used in BERT models to aggregate the meaning of the entire input sequence, making it suitable for classification tasks.

### 3.5.4 Development of Augmented FinBERT

In this research, we enhance the FinBERT model by incorporating augmented data generated via GPT-4o, leading to the creation of Augmented FinBERT. This refined model is designed to more effectively handle the intricacies of financial sentiment analysis by integrating synthetic examples that capture a wider spectrum of financial expressions and subtleties.

**Data Augmentation with GPT-4o**

The augmentation process involves generating synthetic data using GPT-4o that mimics challenging scenarios and boundary cases previously misclassified by FinBERT. This data, combined with the original training corpus, enriches the training set with diverse examples, thereby enhancing the model's ability to generalize across various financial texts. The integration of this augmented data provides a richer learning context, enabling FinBERT to maintain high accuracy while improving its adaptability to nuanced financial sentiments.

**Fine-tuning Strategy**

To ensure robustness and prevent degradation of pre-trained capabilities, we implement advanced fine-tuning strategies as outlined by Howard and Ruder, 2018a. They caution that catastrophic forgetting is a significant risk during the fine-tuning of neural network models, a phenomenon where the model may lose knowledge acquired during its initial training as it adapts to a new task.

To mitigate this risk in the development of Augmented FinBERT, we apply three strategic techniques: slanted triangular learning rates, discriminative fine-tuning, and gradual unfreezing. These methods, also utilized in the initial training of FinBERT as noted by Araci, 2019, are designed to ensure that FinBERT retains its foundational language capabilities while effectively adapting to the specialized demands of financial sentiment analysis.

To fine-tune the Augmented FinBERT effectively while mitigating the risk of catastrophic forgetting, we employ the following three-pronged strategy:

**Discriminative Fine-tuning**

We implement discriminative fine-tuning by assigning lower learning rates to deeper layers of the network. Assuming the learning rate at layer $l$ is $\alpha$, the learning rate for each preceding layer $l-1$ is defined as $\alpha_{l-1} = \theta \alpha_l$ where $\theta$ is the discrimination rate. This technique ensures that lower layers, which capture more generic language features crucial for foundational linguistic understanding, receive smaller updates to preserve their broad applicability. Simultaneously, it allows upper layers, which are more task-specific, to adapt more responsively to the specialized requirements of financial sentiment analysis tasks.

**Parameter Grouping in Discriminative Fine-Tuning**

In addition to applying different learning rates across the network layers, we also employ a nuanced approach to parameter handling within each layer. This involves grouping parameters based on their susceptibility to decay.

**Gradual Unfreezing**

We employ a gradual unfreezing method to meticulously manage the adaptation of the FinBERT model to augmented data while preserving its pre-trained linguistic capabilities. Initially, all layers of the model, except for the classifier, are frozen. This approach is designed to prevent early modifications to the model's fundamental linguistic patterns, which are crucial for general language understanding.

As training progresses, we implement a staged unfreezing process. This begins with the topmost layer (nearest to the output), which is unfrozen first, and subsequently moves downwards to the lower layers. This controlled, incremental exposure of each layer to the training process helps mitigate the risk of catastrophic forgetting, where the model might lose valuable pre-trained knowledge.

This strategy mirrors the successful approach utilized in the original training of FinBERT. By only allowing the classifier to train initially and then progressively unfreezing the remaining layers, we balance the need for task-specific adaptation with the preservation of foundational linguistic knowledge. This careful calibration allows Augmented FinBERT to enhance its performance on complex financial texts without compromising the robustness of its language modeling.

**Slanted Triangular Learning Rates**

This approach applies a learning rate that initially increases and then decreases in a slanted triangular fashion. This means that the learning rate starts at a lower value and gradually increases, allowing the model to adapt smoothly to the new augmented data before reaching the peak learning rate. After this warm-up phase, the learning rate begins to decrease, helping to refine the model's adjustments. This dynamic adjustment of learning rates aids in stabilizing the learning process, ensuring that the model effectively integrates the new augmented data without compromising the knowledge it acquired during pre-training and initial training.

**Conclusion**

The methodology implemented for the development of Augmented FinBERT strategically utilizes state-of-the-art fine-tuning techniques tailored to enhance performance in financial sentiment analysis. By incorporating synthetic data generated by large language models (LLMs), Augmented FinBERT benefits from enriched training sets that simulate complex and nuanced financial scenarios. This integration of knowledge distilled from advanced LLMs through data augmentation significantly boosts the model's ability to interpret and analyse financial sentiments accurately. Consequently, Augmented FinBERT is not only expected to surpass FinBERT in terms of performance but also demonstrate improved robustness and adaptability in handling the intricate demands of financial sentiment analysis. This approach exemplifies how LLMs can be leveraged to refine and extend the capabilities of existing smaller models, ensuring they remain effective in the evolving landscape of financial analysis.

### 3.5.5 Performance Metrics

To assess the effectiveness of Augmented FinBERT and TinyFinBERT, we utilize several key performance metrics: accuracy, precision, recall, and F1 scores. Each of these metrics plays a critical role in evaluating different aspects of the model's classification capabilities:

- **Accuracy:** Measures the overall correctness of the model across all classifications, providing a general sense of its effectiveness.
- **Precision:** Indicates the accuracy of positive predictions, essential for financial applications where false positives can have significant implications.
- **Recall (Sensitivity):** Assesses the model's ability to identify all relevant instances, crucial for not overlooking critical financial sentiments.
- **F1 Score:** Combines precision and recall into a single metric that balances both, vital in conditions where both false positives and false negatives are costly.

These metrics are chosen to address the complexities presented by the imbalanced nature of the Financial PhraseBank dataset, where 59% of the records are labeled as Neutral, 28% as Positive, and only 12% as Negative. Such an imbalance makes it crucial to employ metrics that can accurately reflect the model's performance across unevenly distributed classes. Precision and recall are particularly important in this context, as they help understand how effectively the model handles less-represented classes, which is vital for ensuring robustness and fairness in financial sentiment analysis. The F1 score is equally important as it provides a measure of the model's accuracy that considers both the precision and the recall, which is crucial for dealing with the skewed distribution of classes.

### 3.5.6  Measuring Performance

The performance evaluation of Augmented FinBERT is conducted using two primary methodologies:

**Post-Augmentation Evaluation:**

We assess Augmented FinBERT using a test set specifically reserved from the Financial PhraseBank dataset. This evaluation compares the performance of Augmented FinBERT against the original FinBERT model, allowing us to measure improvements in accuracy, precision, recall, and F1 scores after incorporating augmented data.

**Baseline Comparison:**

Additionally, the performance metrics of Augmented FinBERT are compared with those of the original FinBERT model using the FiQA 2018 Task1 and Forex News Annotated datasets. These datasets were not involved in the training process and are utilized to evaluate the model's generalization capabilities. This comparison is vital for assessing how Augmented FinBERT performs under conditions that mimic real-world financial sentiment analysis, especially its ability to generalize to new and unseen data.

### 3.5.7 Knowledge Distillation Strategy

Knowledge Distillation (KD) is a process used to transfer knowledge from a large, complex teacher network ($T$) to a smaller, simpler student network ($S$). The student network learns by mimicking the behavior of the teacher network. In this context, the behavior functions $f^T$ and $f^S$ of the teacher and student networks respectively, convert network inputs into informative representations. These behavior functions could be outputs from various layers within the network, such as the Multi-Head Attention (MHA) layer, Feed-Forward Network (FFN) layer, or even the attention matrices. The essence of KD is captured by an objective function designed to minimize the loss $L$, which measures the discrepancies between the outputs of the student and teacher networks across all inputs $x$ from the training dataset $\mathcal{X}$:

$$\mathcal{L}_{\text{KD}} = \sum_{x \in \mathcal{X}} L(f^S(x), f^T(x)) \tag{5}$$

This loss function helps quantify how closely the student network is able to replicate the behavior of the teacher network.

**Transformer Distillation Technique for Knowledge Transfer**

For performing Knowledge Distillation we used the same Transformer distillation method used for the TinyBERT model (Jiao et al., 2019). The proposed Transformer distillation method is tailored for distilling knowledge from a larger teacher network (with N Transformer layers) to a smaller student network (with M Transformer layers). The process involves selecting corresponding layers from the teacher network to target during distillation, defining a mapping function $n = g(m)$ where the $m$-th student layer learns from the $g(m)$-th teacher layer.

**Problem Formulation and Objective Function:**

The objective of Transformer distillation is defined as:

$$\mathcal{L}_{\text{model}} = \sum_{x \in \mathcal{X}} \sum_{m=0}^{M+1} \lambda_m \mathcal{L}_{\text{layer}}\left(f_m^S(x), f_{g(m)}^T(x)\right) \tag{6}$$

where $\mathcal{L}_{\text{layer}}$ is the loss function for a specific layer, $f_m(x)$ denotes the behavior function from the $m$-th layer, and $\lambda_m$ is a hyperparameter dictating the importance of each layer's distillation.

**Distillation Components:**

1. **Attention-based Distillation:**

   Focuses on the multi-head attention matrices which encapsulate rich linguistic knowledge. The distillation loss for attention matrices is defined as:

   $$\mathcal{L}_{\text{attn}} = \frac{1}{h} \sum_{i=1}^{h} \text{MSE}\left(A_i^S, A_i^T\right) \tag{7}$$

   where $h$ is the number of attention heads, $A_i$ represents the attention matrix for the $i$-th head, and MSE denotes the mean squared error.

2. **Hidden States-based Distillation:**

   Involves distilling the output of the Transformer layers' hidden states:

   $$\mathcal{L}_{\text{hidn}} = \text{MSE}\left(H^S W_h, H^T\right) \tag{8}$$

   where $H^S$ and $H^T$ are the hidden states of the student and teacher, respectively, and $W_h$ is a learnable transformation matrix aligning student hidden states with those of the teacher.

3. **Embedding-layer Distillation:**

   Similar to hidden states distillation, focusing on the embeddings:

   $$\mathcal{L}_{\text{embd}} = \text{MSE}\left(E^S W_e, E^T\right) \tag{9}$$

   where $E^S$ and $E^T$ refer to the embeddings of student and teacher

4. **Prediction-layer Distillation:**

   Utilizes the soft cross-entropy loss to align the predictions of the student with those of the teacher:

   $$\mathcal{L}_{\text{pred}} = \text{CE}(z^T/t, z^S/t) \tag{10}$$

   where $z^S$ and $z^T$ are the logits from the student and teacher, CE denotes cross entropy loss, and $t$ is the temperature scaling factor.

5. **Unified Distillation Loss:**

   By unifying the distillation objectives outlined above (Equations 7, 8, 9, and 10), we effectively consolidate the distillation loss across the corresponding layers between the teacher and student networks, ensuring a cohesive and targeted knowledge transfer:

$$\mathcal{L}_{\text{layer}} = \begin{cases} \mathcal{L}_{\text{embd}}, & m = 0 \\ \mathcal{L}_{\text{hidn}} + \mathcal{L}_{\text{attn}}, & M \geq m > 0 \\ \mathcal{L}_{\text{pred}}, & m = M + 1 \end{cases} \quad (11)$$

## 3.6 Summary

This chapter has systematically outlined the research methodology utilized in this study, encompassing a broad spectrum of techniques and strategies essential for advancing financial sentiment analysis using large language models. The methodologies described span from the initial design of the research through detailed descriptions of the datasets, synthetic data generation, model development, and performance evaluation.

Key components of the research methodology include:
- **Research Design:** Establishes the framework and sequence of approaches used to integrate and analyse the data, setting the stage for a rigorous investigation.
- **Dataset Description:** Provides comprehensive details about the datasets used in this study, Financial PhraseBank, FiQA 2018 Task1, and Forex News Annotated, each critical for assessing different aspects of the models' performance.
- **Synthetic Data Generation:** Describes innovative strategies for augmenting data using LLMs to create new training examples and variations. This process not only enriches the training data but also ensures efficient use of resources.
- **Model Development and Fine-tuning:** Covers the development processes for BERT, FinBERT, and the enhanced Augmented FinBERT, followed by specific fine-tuning techniques to optimize performance.

- **Performance Metrics:** Explains the metrics employed to evaluate the models, highlighting the importance of accuracy, precision, recall, and F1 scores in financial sentiment analysis.
- **Knowledge Distillation:** Details the strategic transfer of knowledge from Augmented FinBERT to TinyFinBERT, ensuring the smaller model retains the capability to perform complex financial sentiment analysis tasks.

The methodologies employed are designed to leverage the cutting-edge capabilities of LLMs like GPT-4 Omni to enhance smaller financial sentiment analysis models.

# CHAPTER 4 : ANALYSIS

## 4.1 Introduction

This chapter delves into the comprehensive analysis undertaken as part of this research, pivotal in assessing the robustness and effectiveness of the methodologies employed. The analysis spans multiple dimensions, from exploratory data analysis (EDA) of the datasets utilized, through detailed model parameter examination, to the evaluation of synthetic data generated for training and augmenting the models. Each section is structured to both detail the findings and also contextualize them within the broader goals of this study.

The exploratory data analysis offers insights into the core datasets, Financial PhraseBank, FiQA 2018 Task1, and Forex News Annotated. This analysis aids in understanding the data characteristics that could influence model training and performance.

Subsequent sections focus on model parameter analysis for FinBERT, Augmented FinBERT, TinyBERT, and TinyFinBERT. This includes a focused review of the models created in this study (Augmented FinBERT and TinyFinBERT), dissecting their structure and behavior under different configurations, providing a deeper understanding of how architectural and parameter choices affect performance.

The chapter progresses to examine the synthetic data generated through Large Language Models (LLMs). This section provides details on the new synthetic training sentences and variations of mislabelled sentences.

## 4.2 Exploratory Data Analysis (EDA)

### 4.2.1 Financial PhraseBank Dataset

The Financial PhraseBank dataset, developed by Malo et al. (2014), comprises 4,846 sentences from financial news texts and company press releases, sourced from the LexisNexis database. For this study, the dataset used includes all records that have at least 50% agreement among reviewers, consistent with the methodology used in the FinBERT model. This approach allows for a fair comparison and enhances data variability and robustness by employing a merged dataset from various levels of agreement. The specific file used, "Sentences_50Agree.txt," contains 4,846 records, with these categorizations further detailed in Table 4.1, which only displays unique records for each level of agreement.

Table 4.1: Distribution of sentiment labels and agreement levels in PhraseBank data

| Agreement Level | Negative | Neutral | Positive | Total |
|---|---|---|---|---|
| 50% - 65% Agreement | 2% | 7% | 4% | 13% |
| 66% - 74% Agreement | 2% | 8% | 6% | 16% |
| 75% - 99% Agreement | 2% | 16% | 7% | 25% |
| 100% Agreement | 6% | 29% | 12% | 47% |
| **Total** | **12%** | **59%** | **28%** | **100%** |

As can be seen in Table 4.1, around 47% of the data comprises of records with 100% agreement while only 13% of the records have agreement level between 50% and 65%. More than half (59%) of the records have Neutral sentiment, followed by Positive sentiment (28%), and Negative sentiment (12%). These factors have been considered during data augmentation.

For the test part of the data we get the following

Table 4.2: Distribution of sentiment labels and agreement levels in PhraseBank test data

| Label | Count |
|---|---|
| neutral | 575 |
| positive | 267 |
| negative | 128 |

Additionally, an assessment of the sentences in Phrase bank dataset shows that almost 99% of the sentences are under 64 tokens thus supporting the decision to use max token length as 64 in our study.

### 4.2.2 FiQA 2018 Task1 Dataset

The FiQA 2018 Task1 dataset comprises 438 financial headlines and 675 financial tweets, totaling 1,113 records in the training set. This diverse collection, sourced from platforms such as StockTwit and Reddit, covers a broad spectrum of financial topics. Each entry is annotated with a continuous sentiment score ranging from -1 (negative) to +1 (positive), providing a rich resource for testing sentiment analysis models.

For the purposes of this study, these continuous scores have been categorized into three discrete classes to allow for direct comparisons with other sentiment analysis models and the Financial PhraseBank dataset:

- **Negative:** Scores less than -0.5.
- **Neutral:** Scores between -0.5 and 0.5.
- **Positive:** Scores greater than 0.5.

This re-categorization ensures consistency in analysis across different datasets. Following this conversion, the distribution of sentiments in the FiQA 2018 Task1 dataset is as follows:

Table 4.3: Distribution of sentiment labels and agreement levels in FiQA 2018 Task1 Dataset

| Label | Count |
|---|---|
| neutral | 823 |
| positive | 187 |
| negative | 103 |

These figures facilitate direct comparisons between the outcomes derived from the FiQA 2018 Task1 and those from the Financial PhraseBank dataset, enhancing the robustness and validity of the research findings.

The sentiment distribution in the FiQA 2018 Task1 dataset exhibits a pattern similar to that of the Financial PhraseBank dataset, with a majority of records categorized as neutral. This predominance of neutral sentiments, representing approximately 74% of the FiQA dataset, mirrors the distribution in the Financial PhraseBank, where neutral sentiments also prevail. Such similarity in distribution patterns is particularly beneficial for this study as it provides a consistent basis for evaluating the performance of sentiment analysis models across varied datasets. This uniformity ensures that the findings are not only robust but also applicable across different sources of financial textual data, thereby enhancing the generalizability of the research results.

### 4.2.3 Forex News Annotated dataset

The Forex News Annotated dataset, created by (Fatouros et al., 2023), consists of 2,291 financial news headlines from the year 2023, specifically focusing on major forex pairs such as AUDUSD, EURCHF, EURUSD, GBPUSD, and USDJPY.

Table 4.4: Distribution of sentiment labels and agreement levels in Forex News Annotated Dataset

| Label | Count |
|---|---|
| neutral | 815 |
| positive | 767 |
| negative | 709 |

Based on the results in Table 4.4 it can be seen that the distribution of sentiments within the dataset is relatively balanced, enhancing the dataset's utility for training and evaluating sentiment analysis models. Specifically, the dataset contains 815 neutral headlines, making it the most common sentiment, closely followed by 767 positive and 709 negative headlines. This balanced distribution is beneficial for machine learning models as it provides ample examples across all sentiment categories, reducing potential biases towards any single sentiment and promoting better generalization in model performance.

## 4.3    Model Parameter Analysis

### 4.3.1    FinBERT

FinBERT is hosted on Hugging Face[3] under "*ProsusAI/finbert*". Leveraging the robust architecture of BERT-base, FinBERT incorporates 12 encoder layers, a hidden size of 768, 12 multi-head attention heads, and approximately 110 million parameters, ensuring a deep and nuanced understanding of complex financial texts. A dense layer to process the output from the [CLS] token is added, this is traditionally used in BERT models to aggregate the meaning of the entire input sequence, making it suitable for classification tasks.

### 4.3.2    Augmented FinBERT

Similar to FinBERT, Augmented FinBERT has 12 Transformer layers, with a hidden size of 768, feedforward size of 3072, and 12 attention heads, summing up to 110 million parameters. As outlined in Section 3.5.4, the fine-tuning of Augmented FinBERT incorporates three strategic techniques: slanted triangular learning rates, discriminative fine-tuning, and gradual unfreezing.

**Discriminative Fine-tuning:** During the initial training of FinBERT, a discrimination rate of 0.85 was employed. However, for the subsequent fine-tuning that incorporates augmented data for developing Augmented FinBERT, we have adjusted this to a discrimination rate of 0.95. This modification is designed to slightly decelerate the reduction of learning rates across the model's layers. By doing so, it provides a more conservative approach to updating the pre-trained weights. This decision stems from the understanding that the model, already somewhat optimized, benefits from a less steep reduction in learning rates. This approach aids in preserving the refined linguistic nuances previously learned, while still accommodating the integration of new and complex financial terminologies found in the augmented dataset.

---

[3] https://huggingface.co/ProsusAI/finbert

**Parameter Grouping in Discriminative Fine-Tuning**

we also employ a nuanced approach to parameter handling within each layer. This involves grouping parameters based on their susceptibility to decay:

- **Differentiated Decay:** Parameters within each layer are grouped based on whether they are prone to decay (such as weights) or not (such as biases and normalization parameters). This differentiation allows us to apply a weight decay only to those parameters that benefit from it, thereby enhancing training stability.
- **No Decay for Certain Parameters:** Essential parameters like biases and LayerNorm weights do not undergo decay, maintaining their stability across training iterations. This helps preserve the integrity of learned features while allowing for necessary adjustments to adapt to new data.

**Gradual Unfreezing**

Initially, all layers of the model, except for the classifier, are frozen. We implement a staged unfreezing process where each layer is sequentially unfrozen after every third of a training epoch.

**Slanted Triangular Learning Rates**

This approach applies a learning rate that initially increases and then decreases in a slanted triangular fashion, incorporating a warm-up proportion of 0.2. This means that the learning rate starts at a lower value and gradually increases during the initial 20% of the training period, allowing the model to adapt smoothly to the new augmented data before reaching the peak learning rate.

**Incorporating Dropout to Enhance Model Generalization**

In the training of Augmented FinBERT, we incorporate a dropout rate of 0.1 as part of our strategy to prevent overfitting. This dropout probability is applied across the network's layers during training to randomly deactivate a subset of neurons, effectively thinning out the network temporarily. By doing so, we reduce the chance of dependency on any single or small group of neurons during training, promoting a more robust and generalized learning across different parts of the network. This helps ensure that the model remains effective and reliable when analyzing diverse financial texts, enhancing its ability to generalize from training data to real-world applications.

**Training Parameters and Model Configuration**

Consistent with the training specifications of FinBERT, we also configure Augmented FinBERT with a maximum sequence length of 64 tokens. This sequence length is optimal for capturing the necessary context in financial statements without introducing excessive padding or truncation. We set the learning rate at 2e-5 and use a mini-batch size of 64 to balance the trade-off between computational efficiency and model performance. The model is trained over 6 epochs to ensure sufficient exposure to the augmented data while preventing overfitting. During training, performance is continuously monitored on a validation set, and the model that exhibits the best performance is selected for final evaluation. This approach ensures that Augmented FinBERT is finely tuned to deliver optimal accuracy in real-world financial sentiment analysis tasks.

### 4.3.3   TinyBERT

TinyFinBERT utilizes the "*huawei-noah/TinyBERT_General_4L_312D*" model pre-trained on a broad corpus, hosted on Hugging Face[4]. This model, comprising 4 Transformer layers, a hidden size of 312, a feedforward size of 1200, and 12 attention heads, totalling 14.5 million parameters.

### 4.3.4   TinyFinBERT

TinyFinBERT has demonstrated robust general language capabilities (Jiao et al., 2019). Based on findings from (Araci, 2019), which suggested minimal impact of pre-training on Financial Corpus on FinBERT's task-specific performance, our distillation process focuses exclusively on task-specific distillation where we only use the Financial Phrasebank dataset along with augmented unlabelled data created using GPT-3.5T and GPT-4o. Augmented FinBERT is used as the Teacher Model. Similar to TinyBERT, TinyFinBERT has the same attributes.  4 Transformer layers, a hidden size of 312, a feedforward size of 1200, and 12 attention heads, totalling 14.5 million parameters

---

[4] https://huggingface.co/huawei-noah/TinyBERT_General_4L_312D

**Layer Mapping and Distillation Strategy**

Similar to the approach used in development of TinyBERT we apply a layer mapping function $g(m) = 3 \times m$, chosen to maximize coverage of the teacher model's capabilities with fewer student layers, facilitating comprehensive learning across spaced intervals. This ensures TinyFinBERT captures essential features despite its reduced scale.

**Distillation Objectives and Parameters:**

Distillation is performed at a temperature of 1, reflecting direct probability transfer, which is optimal for maintaining fidelity to the teacher model's outputs. Each corresponding layer's contribution is weighted equally ($\lambda = 1$), simplifying the loss computation and focusing on effective, balanced knowledge transfer. During distillation unlabelled augmented data generated by GPT 3.5T and GPT 4o along with the original training data is utilized.

**Distillation Phases**

The distillation process for TinyFinBERT is meticulously structured in two phases, closely following the strategies implemented in the original TinyBERT framework:

**Intermediate Layer Distillation:**

Initially, we perform intermediate layer distillation, which targets the deeper, non-output layers of the networks. This phase is conducted over 20 epochs using a batch size of 32 and a learning rate of $5e - 5$. The maximum sequence length during this training is set to 64 tokens, and a warm-up proportion of 0.1 is applied to gradually ramp up the learning rate at the beginning of the training process. This phase focuses on adapting the student model's intermediate representations to closely mirror those of the teacher model, leveraging the LLM augmented data to enrich the training process.

**Prediction Layer Distillation:**

Following the initial phase, prediction layer distillation is performed for 3 epochs, with all other parameters having the same values used in intermediate layer distillation. This phase concentrates on the output layer of TinyFinBERT, where the model learns to replicate the final decision-making process of the Augmented FinBERT. The short duration of this phase reflects its focused objective of refining the topmost layer's ability to predict based on the refined intermediate features.

**Comprehensive Approach:** By splitting the distillation into two distinct phases, we ensure that TinyFinBERT not only learns the robust features from Augmented FinBERT's deeper layers but also fine-tunes its output predictions to align closely with those of the teacher model. This structured approach optimizes the transfer of nuanced understanding from Augmented FinBERT, enhancing TinyFinBERT's performance on complex financial sentiment analysis tasks.

## 4.4 Synthetic Data Analysis

### 4.4.1 Using LLM to Generate New Training Examples

Following the methodology outlined in Section 3.4.1, GPT-4 Omni was employed to generate new training examples. Initially, 693 sentences were created using a variety of prompts. The sentiment distribution of these generated sentences is presented in Table 4.5.

Table 4.5: Distribution of sentiment labels for GPT-4o generated data

| Label | Count |
|---|---|
| neutral | 400 |
| positive | 149 |
| negative | 144 |

In accordance with the methodology detailed in Section 3.4.2, these synthetic examples underwent a labelling process using GPT-4 Omni, which was repeated five times with the data order randomly shuffled in each iteration. This rigorous labelling exercise resulted in 410 labelled sentences suitable for fine-tuning. The sentiment distribution of these filtered sentences is shown in Table 4.6.

Table 4.6: Distribution of sentiment labels for filtered GPT-4o generated data used in finetuning

| Label | Count |
|---|---|
| neutral | 377 |
| positive | 28 |
| negative | 5 |

It is evident that the majority of the discarded sentences were initially generated to convey positive and negative sentiments but were subsequently found to have different sentiments upon labelling with GPT-4 Omni. These mismatches led to their exclusion from the fine-tuning dataset to ensure sentiment accuracy and consistency.

### 4.4.2 Using LLM to Create Variations of Mislabeled sentences

Following the methodology outlined in Section 3.4.3, we first identified sentences in the training data that were mislabelled by FinBERT. Out of a total of 3,488 sentences in the training dataset, 382 sentences were found to be mislabelled. The sentiment distribution of these mislabelled sentences is presented in Table 4.7.

Table 4.7: Distribution of sentiment labels for sentences mislabelled by FinBERT

| Label | Count |
| --- | --- |
| neutral | 301 |
| positive | 70 |
| negative | 11 |

In the next step, these sentences were relabeled using GPT-4 Omni through a stringent five-step iteration process, as described in the methodology for initial labelling. Only sentences where GPT-4 Omni's labelling matched the original dataset labels were retained. This filtering resulted in 124 sentences. The sentiment distribution of these remaining sentences used for GPT-4 Omni augmentation is shown in Table 4.8.

Table 4.8: Distribution of sentiment labels for training sentences used for GPT-4o augmentation

| Label | Count |
| --- | --- |
| neutral | 61 |
| positive | 53 |
| negative | 10 |

Interestingly, most of the discarded sentences were those labeled as neutral in the Financial PhraseBank dataset. GPT-4 Omni, with its advanced understanding, may account for factors that the original labelers might have missed or that were not part of the original labelling guidelines. Alternatively, GPT-4 Omni could also be mislabeling these sentences. To ensure the augmented data generated by GPT-4 Omni is consistent with the Financial PhraseBank dataset, we use this relabeling step to weed out instances of conflict between GPT-4 Omni and the original labelers. This ensures the integrity and consistency of the augmented data for fine-tuning purposes.

In the next step, GPT-4 Omni was tasked with generating 10 variations for each of the remaining 124 sentences. This process resulted in a total of 1,219 statements. In some instances, GPT-4 Omni generated fewer than 10 variations per sentence, likely due to the frequency penalty being set to 0.5. For a detailed discussion of the API parameters used in this step, please refer to Section 3.4.3. Consequently, instead of the expected 1,240 variations (124 sentences x 10 variations each), 1,219 sentences were ultimately generated.

Following this, these sentences were relabeled using GPT-4 Omni through a rigorous five-iteration process, as described in the methodology for initial labelling. Only sentences where GPT-4 Omni's labelling matched the original dataset labels were retained. This filtering resulted in 1,001 sentences. The sentiment distribution of these augmented sentences is presented in Table 4.9.

Table 4.9: Distribution of sentiment labels for GPT-4o augmented variations of existing sentences

| Label | Count |
|---|---|
| neutral | 505 |
| positive | 421 |
| negative | 75 |

These 1,001 GPT-4o generated variations of existing sentences along with 410 sentences generated purely using GPT 4o prompting comprised the augmented data used for finetuning Augmented FinBERT.

### 4.4.3 Using LLM to Generate New Unlabeled Examples

In the process of creating variations, we generate 30 new variations for each piece of FinBERT mislabelled data and 5 new variations for each correctly labeled data in the Financial PhraseBank training dataset. The decision to generate 30 new variations for each piece of FinBERT mislabelled data and 5 new variations for correctly labeled data is strategically designed to optimize the knowledge distillation process. Mislabelled data inherently present more significant learning opportunities and challenges; thus, generating a higher number of variations for these examples helps the student model (TinyFinBERT) to better learn from complex or ambiguous cases. Conversely, correctly labeled data are already well-understood by the model, requiring fewer variations to reinforce correct sentiment detection. This differential approach ensures efficient use of computational resources while maximizing the learning potential from more problematic data points, thereby enhancing the overall robustness and accuracy of TinyFinBERT.

There are 3,494 sentences correctly labeled by FinBERT. The sentiment distribution of these correctly labeled sentences is presented in Table 4.10.

Table 4.10: Distribution of sentiment labels for sentences correctly labelled by FinBERT

| Label | Count |
|---|---|
| neutral | 2003 |
| positive | 1026 |
| negative | 465 |

Table 4.7 above shows the sentiment distribution of the 382 mislabelled sentences. Following GPT-3.5T prompting, 17,458 sentences were generated for the 3,494 correctly labeled sentences, and 13,148 sentences were generated for the 382 mislabelled sentences. Additionally, 3,028 GPT-4o-generated sentences that were not used in earlier steps are also included as unlabeled data. The total number of unlabeled sentences, therefore, comes to 33,634. These unlabeled sentences are then used for performing knowledge distillation on TinyFinBERT.

## 4.5 Summary

This chapter has explored various analytical dimensions integral to validating the efficacy and robustness of the enhanced sentiment analysis models developed in this research. Through the exploratory data analysis (EDA) of the Financial PhraseBank, FiQA 2018 Task1, and Forex News Annotated datasets, we gained crucial insights into the data characteristics that directly influence model performance and outcomes. This foundational analysis ensures that subsequent modeling decisions are well-informed and tailored to the nuances of financial sentiment detection.

In-depth parameter analysis of FinBERT, Augmented FinBERT, TinyBERT, and TinyFinBERT has provided a deeper understanding of each model's capabilities and limitations. These insights are critical in fine-tuning the models to achieve optimal performance.

Furthermore, the evaluation of synthetic data generated through the use of LLMs has illustrated the significant role of augmented data in enhancing model training. By generating new training examples and variations of mislabelled sentences, we have effectively expanded the training corpus, allowing for more robust learning and generalization across diverse financial texts.

Overall, the analyses conducted in this chapter underscore the effectiveness of the advanced methodologies employed in this study. The findings not only validate the approaches used but also contribute valuable insights to the ongoing development and refinement of sentiment analysis models within the financial sector.

# CHAPTER 5 : RESULTS AND DISCUSSIONS

## 5.1 Introduction

This chapter presents a comprehensive analysis of the performance outcomes from the study, discussing the effects and implications of integrating LLM data augmentation and knowledge distillation techniques into financial sentiment analysis models. Specifically, it delves into the performance metrics of the Augmented FinBERT compared to the baseline FinBERT model across multiple datasets including the Financial PhraseBank, FIQA 2018 Task1, and the Forex News Annotated datasets. Subsequent sections explore the results of knowledge distillation applied to TinyFinBERT and evaluate the efficacy of the transferred knowledge from the advanced LLMs. The discussions aim to not only quantify the improvements but also to interpret the underlying factors contributing to these results, providing a critical analysis of the models' behavior in various testing scenarios. This synthesis of results leads to insights that form the basis for proposed future directions in enhancing model performance and exploring new applications.

## 5.2 Performance of Augmented FinBERT

The performance of the Augmented FinBERT model, which was fine-tuned using synthetic data generated from advanced Large Language Models (LLMs), demonstrates significant improvements over the baseline FinBERT model. The enhancements are evident across all metrics, accuracy, F1 score, precision, and recall, as presented in Table 5.1.

Table 5.11: Performance results for Augmented FinBERT

| Model | Dataset | Accuracy | F1 Score | Precision | Recall |
|---|---|---|---|---|---|
| FinBERT | FPB Test* | 0.8423 | 0.8439 | 0.8545 | 0.8423 |
|  | Forex | 0.4801 | 0.4449 | 0.4988 | 0.4801 |
|  | FIQA | 0.5265 | 0.5563 | 0.6642 | 0.5265 |
| Augmented FinBERT | Test | **0.8742** | **0.8739** | **0.8743** | **0.8742** |
|  | Forex | **0.495** | **0.4797** | **0.5081** | **0.495** |
|  | FIQA | **0.6217** | **0.6385** | **0.6709** | **0.6217** |

### 5.2.1 Financial PhraseBank Test Dataset (FPB Test)

**Performance Enhancement:** Augmented FinBERT demonstrated marked improvements, achieving an accuracy increase from 0.8423 to 0.8742 and an F1 score boost from 0.8439 to 0.8739.

This enhancement highlights the benefits of integrating synthetic data, which exposes the model to a wider range of financial sentiments and linguistic nuances, thereby increasing its adaptability and overall accuracy.

**Comparison with Original FinBERT Study:** It's important to note that while the original FinBERT study reported(Araci, 2019) an accuracy of 0.86 and an F1 score of 0.84 on the test data, our replication efforts yielded slightly different results. Using the FinBERT model available on Hugging Face, coupled with the author-provided train-test split logic, we observed an accuracy of 0.84, slightly below the reported 0.86. Despite this, the accuracy of Augmented FinBERT at 0.87 surpasses both the replicated and originally reported accuracies, underscoring the effectiveness of the augmented training approach.

### 5.2.2 FIQA 2018 Task1 Dataset

The substantial gains on the FIQA dataset, with accuracy rising from 0.5265 to 0.6217 and F1 score from 0.5563 to 0.6385, demonstrate the model's enhanced capability to understand and analyse financial opinions more accurately. Enhanced performance on this unseen dataset indicates that the incorporation of LLM-augmented data effectively prevented overfitting, while simultaneously boosting accuracy on the Financial PhraseBank dataset. This is further evidenced by the notable improvements in generalization capabilities, as demonstrated by the increased performance metrics observed on the FIQA dataset. This dataset's results are particularly significant as they reflect the model's improved performance in a complex financial sentiment analysis task, aligning with the thesis aim to enhance model robustness and precision.

### 5.2.3 Forex News Annotated Dataset

Although the improvement in Forex dataset performance is modest (from 0.4801 to 0.4950 in accuracy), it highlights the challenges associated with adapting models to highly volatile and specialized domains like Forex markets. The slight increase in precision and F1 score indicates that Augmented FinBERT is better at identifying relevant sentiments, which is crucial for real-time financial analysis.

### 5.2.4 Conclusion

The objectives of this thesis focused on leveraging the generative capabilities of LLMs like GPT-4 Omni to enrich the training dataset and thus improve the model's performance in specialized financial sentiment analysis tasks. The augmented data helped address data scarcity issues and introduced a broader spectrum of financial terminologies and contexts, which was crucial for the fine-tuning process.

The comparative analysis with the baseline FinBERT model reveals that the strategic incorporation of synthetic data not only enhances model performance across traditional metrics but also ensures that the model is better equipped to handle real-world financial texts, thereby increasing its practical applicability and reliability in dynamic financial environments.

Overall, the results validate the hypothesis that augmenting the training dataset with high-quality, diverse synthetic examples can significantly improve the effectiveness of BERT based sentiment analysis models. This approach, as demonstrated by the performance of Augmented FinBERT, not only advances the field of NLP in finance but also sets a benchmark for future research in the application of LLM augmented data in enhancing financial sentiment classification tasks.

## 5.3 Knowledge Distillation Outcomes

The development of TinyFinBERT aimed to harness the capabilities of Augmented FinBERT while reducing the model size, thus making it more feasible for deployment in resource-constrained environments. This section analyzes TinyFinBERT's performance compared to both the baseline FinBERT and the intermediary Augmented FinBERT across three distinct datasets.

Table 5.12: Performance results for TinyFinBERT

| Model | Dataset | Accuracy | F1 Score | Precision | Recall |
|---|---|---|---|---|---|
| **FinBERT** | FPB Test | 0.8423 | 0.8439 | 0.8545 | 0.8423 |
| | Forex | 0.4801 | 0.4449 | 0.4988 | 0.4801 |
| | FIQA | 0.5265 | 0.5563 | 0.6642 | 0.5265 |
| Augmented FinBERT | Test | **0.8742** | **0.8739** | **0.8743** | **0.8742** |
| | Forex | **0.495** | **0.4797** | **0.5081** | **0.495** |
| | FIQA | **0.6217** | **0.6385** | **0.6709** | **0.6217** |
| TinyBERT | Test | 0.133 | 0.0329 | 0.6103 | 0.133 |
| | Forex | 0.3095 | 0.1463 | 0.0958 | 0.3095 |
| | FIQA | 0.0925 | 0.0157 | 0.0086 | 0.0925 |
| TinyFinBERT | Test | 0.833 | 0.833 | 0.8333 | 0.833 |
| | Forex | 0.4775 | 0.4572 | 0.4923 | 0.4775 |
| | FIQA | 0.566 | 0.5944 | 0.656 | 0.566 |

Table 5.13: Comparison of TinyFinBERT performance with FinBERT

| Model | Dataset | Accuracy | F1 Score | Precision | Recall |
|---|---|---|---|---|---|
| FinBERT | FPB Test | 0.8423 | 0.8439 | 0.8545 | 0.8423 |
| | Forex | 0.4801 | 0.4449 | 0.4988 | 0.4801 |
| | FIQA | 0.5265 | 0.5563 | 0.6642 | 0.5265 |
| TinyFinBERT | Test | 0.833 | 0.833 | 0.8333 | 0.833 |
| | Forex | 0.4775 | 0.4572 | 0.4923 | 0.4775 |
| | FIQA | 0.566 | 0.5944 | 0.656 | 0.566 |
| TinyFinBERT (as % of FinBERT) | Test | 98.90% | 98.71% | 97.52% | 98.90% |
| | Forex | 99.46% | 102.76% | 98.70% | 99.46% |
| | FIQA | 107.50% | 106.85% | 98.77% | 107.50% |

### 5.3.1 Financial PhraseBank Test Dataset (FPBTest):

As per Table 5.2, TinyFinBERT demonstrates a significant performance boost compared to TinyBERT, nearing the effectiveness of Augmented FinBERT with a dramatic rise in accuracy and F1 scores from 0.133 to 0.833. As detailed in Table 5.3, TinyFinBERT achieves at least 98% of FinBERT's performance across all metrics while being significantly smaller (7.5 times), underscoring the efficiency of the knowledge distillation techniques employed. This achievement aligns with the thesis objective of maintaining high model performance with reduced resource utilization.

### 5.3.2 FIQA 2018 Task1 Dataset:

The results on the FIQA dataset reveal notable improvements in TinyFinBERT's performance over TinyBERT. While TinyFinBERT has not yet reached the performance levels of Augmented FinBERT, the significant gains confirm the effectiveness of the knowledge distillation process in enhancing the model's ability to parse complex financial sentiments. Remarkably, as detailed in Table 5.3, TinyFinBERT not only surpasses the performance of the original FinBERT but also demonstrates that the incorporation of LLM-augmented data has effectively enhanced its generalization capabilities. This improvement on a previously unseen dataset suggests that the use of LLM-augmented data has not only prevented overfitting but has also contributed to higher accuracy on the Financial PhraseBank dataset by transferring some of the LLM knowledge to the Augmented Finbert through Data Augmentation. These advancements in generalization and knowledge transfer from LLM are clearly reflected by the substantial performance increases observed on the FIQA dataset.

### 5.3.3 Forex News Annotated Dataset:

On the Forex dataset, TinyFinBERT significantly outperforms TinyBERT, demonstrating a successful transfer of nuanced financial sentiment detection capabilities from the larger model, despite TinyBERT's initial shortcomings. Although TinyFinBERT's performance is slightly lower than that of Augmented FinBERT, it underscores the inherent challenges of fully encapsulating the complexities of Forex market sentiments within a smaller model framework. Notably, TinyFinBERT achieves nearly comparable results to FinBERT, as seen in Table 5.3, validating the effectiveness of knowledge distillation in maintaining substantial model performance while operating within a more constrained model architecture.

### 5.3.4 Conclusion

These results substantiate the thesis's core objective of employing LLMs for generating synthetic training data and employing them in knowledge distillation techniques to develop a compact yet efficient model like TinyFinBERT. The comparative performance analysis across three datasets illustrates that while TinyFinBERT may not always reach the heights of its larger predecessors, it significantly closes the gap, thus supporting its use in environments where computational efficiency is critical. The enhanced performance of TinyFinBERT across varied financial datasets confirms the viability of using LLM augmented data in knowledge distillation is effective in transferring knowledge from both the larger model and LLM to the smaller, more deployable ones. This approach not only aligns with the aims of reducing computational demands but also maintains the necessary accuracy and robustness required for real-world financial sentiment analysis.

## 5.4 Transfer of Knowledge from LLM

The notable enhancements in the performance of both Augmented FinBERT and TinyFinBERT can be directly attributed to the advanced knowledge transferred from GPT-4 Omni through LLM data augmentation. These improvements are particularly evident in the results on independent datasets such as FIQA Task 1 and the Forex dataset, underscoring the impactful role of LLM data augmentation. The most compelling demonstration of this effect is the far smaller TinyFinBERT's ability to outperform FinBERT on the FIQA Task 1 dataset, showcasing the significant benefits of integrating LLM-generated insights into the training process.

Additionally, Table 5.4 illustrates the performance of LLMs when tasked with labelling the entire Financial PhraseBank test dataset, employing the methodology described in Section 3.4.2, "LLM Labeling of Generated Examples." The results demonstrate that while the accuracy of LLMs such as GPT-3.5 and GPT-4 Omni does not match that of highly specialized models like FinBERT, their performance is commendable given their generalist training across diverse domains.

Table 5.14: LLM Labelling performance on Financial PhraseBank Test Data

| Model | Dataset | Accuracy | F1 Score | Precision | Recall |
|---|---|---|---|---|---|
| **gpt-3.5-turbo-0125** | FPBTest | 0.7314 | 0.7366 | 0.7614 | 0.7314 |
| **gpt-4o-2024-05-13** | FPBTest | 0.7438 | 0.7468 | 0.8043 | 0.7438 |

It's important to note that despite their slightly lower performance metrics compared to specialized models, LLMs possess an expansive knowledge base derived from training on extensive and varied corpora. This broad exposure enables them to offer valuable insights and understanding, which can be leveraged to enhance the performance of smaller, more specialized models. Specifically, the richness of linguistic and semantic nuances captured by LLMs can be distilled into specialized models through strategic data augmentation and distillation processes, potentially reducing the gap in performance metrics and increasing the smaller models' robustness and adaptability to specialized tasks.

This use of LLMs underscores a critical advantage: the ability to generalize and transfer learning from large-scale data to specific domains, which is essential for advancing the capabilities of NLP models in financial sentiment analysis. By harnessing the power of LLMs, researchers can further refine the accuracy and effectiveness of domain-specific models, thus contributing significantly to the field of NLP.

## 5.5    Summary

This chapter provided a detailed exploration of the outcomes from the implementation of advanced training techniques and the utilization of large language models (LLMs) for data augmentation in financial sentiment analysis. We examined the performance enhancements achieved by Augmented FinBERT across different datasets, demonstrating notable improvements in accuracy, precision, recall, and F1 scores compared to the baseline FinBERT model. The use of augmented data from GPT-4 Omni has proven effective, particularly in enhancing model robustness and adaptability to diverse financial linguistic contexts as evidenced in the Financial PhraseBank and FIQA 2018 Task1 datasets.

Furthermore, the results from the knowledge distillation process into TinyFinBERT highlighted the potential of using a smaller, more efficient model without significantly compromising performance. TinyFinBERT's ability to approach, and in some cases surpass, the performance metrics of its larger counterparts underscores the efficacy of the distillation strategies employed, which were carefully designed to transfer nuanced understanding from both LLM and the augmented FinBERT. The discussions also shed light on the transfer of knowledge from LLMs, which contributed significantly to both model enhancement and the broader application potential in real-world scenarios.

By synthesizing these findings, this chapter not only confirms the hypotheses posited at the outset of this research but also sets a foundational framework for future explorations aimed at refining and expanding the use of LLM-augmented models in financial sentiment analysis tasks.

# CHAPTER 6 : CONCLUSIONS AND RECOMMENDATIONS

## 6.1 Introduction

This chapter encapsulates the conclusive insights and recommendations derived from the extensive research conducted on enhancing financial sentiment analysis models using LLMs for data augmentation and knowledge distillation. It synthesizes the findings from the experiments and discussions presented in previous chapters, reaffirming how the strategic incorporation of LLM-augmented data has influenced the performance and generalization capabilities of FinBERT and TinyFinBERT. This chapter is structured to first revisit the study's objectives, discussing how each has been addressed through various methodological applications and experimental outcomes. It then highlights the distinct contributions made to the field of Natural Language Processing (NLP) and financial analytics, proposing pathways for future research to build upon the groundwork laid by this thesis. Following a structured approach, this chapter ensures a comprehensive closure to the study, providing critical reflections on the implications of the findings and suggesting directions for subsequent research endeavors in the domain.

## 6.2 Discussion and Conclusion

**Objective 1: Enhance FinBERT Using LLM-Augmented Data**

- **Methodology Summary:** The approach taken to enhance FinBERT involved generating synthetic data using GPT-4 Omni and GPT-3.5 Turbo, which was then used to fine-tune and develop Augmented FinBERT. This process was designed to address the scarcity of labeled financial sentiment data and to introduce more complex and nuanced examples into the training set.

- **Results and Conclusions:** Augmented FinBERT demonstrated improved performance metrics compared to the original FinBERT model across all evaluated datasets, with notable increases in accuracy and F1 scores on the Financial PhraseBank Test dataset. The performance uplift confirms that LLM-augmented data effectively enriches the training set, leading to a more robust and accurate model.

- **Literature Insights:** The success of this augmentation aligns with findings from the literature that highlight the benefits of synthetic data in overcoming data limitations and enhancing model generalization. Studies like those by (He et al., 2023) provided foundational insights that guided the strategic use of LLMs for targeted data augmentation, confirming that enhanced training data diversity correlates with better model performance.

**Objective 2: Knowledge Distillation to TinyFinBERT**

- **Methodologies Summary:** Knowledge distillation was employed to transfer the learned complexities from Augmented FinBERT to TinyFinBERT. This involved using both intermediate representations and logits to train a smaller, more efficient model capable of maintaining a high level of performance.
- **Results and Conclusions:** TinyFinBERT, despite its reduced size, achieved close to or surpassed the performance metrics of the original FinBERT on certain tasks, particularly on the FIQA dataset. This indicates that significant knowledge was successfully distilled from both Augmented FinBERT and LLMs like GPT-4 Omni and GPT 3.5 Turbo.
- **Comparison with Literature:** Previous models and techniques in knowledge distillation, such as those described by (Hinton et al., 2015), typically showed a trade-off between model size and performance. In contrast, TinyFinBERT's results suggest that with advanced distillation techniques and augmented training data, it is possible to minimize this trade-off, as demonstrated by its comparable performance. This is also in line with recent research on Knowledge Distillation of LLMs (Xu et al., 2024b).

**Objective 3: Evaluation of Generalization Capabilities**

- **Methodology Summary:** The generalization capabilities of Augmented FinBERT and TinyFinBERT were evaluated using datasets that were not involved in the training process, specifically the FIQA 2018 Task1 and Forex News Annotated datasets.
- **Results and Conclusions**: Both models displayed enhanced performance on these datasets compared to their performance on the Financial PhraseBank, with TinyFinBERT showing remarkable improvement over the baseline models. This was particularly evident in its ability to handle the diverse financial sentiments and complex expressions found in the FIQA dataset.

- **Generalization Impact**: The use of LLM-augmented data proved critical in extending the models' generalization capabilities. The improvement across unfamiliar datasets validates the effectiveness of LLM augmentation in enhancing not just the accuracy but also the adaptability of financial sentiment analysis models to new and varied data, suggesting robustness that was previously unattainable with traditional training approaches.

## 6.3 Contribution to Knowledge

This thesis has made several significant contributions to the field of natural language processing and financial sentiment analysis, particularly through the innovative use of large language models for data augmentation and the application of knowledge distillation techniques. The findings from this research not only advance the theoretical understanding of these techniques but also offer practical insights for their application in financial analytics.

1. **Advancement in LLM Data Augmentation**: One of the novel findings of this research is the demonstrable efficacy of LLM-generated synthetic data in enhancing the performance of sentiment analysis models. By employing GPT-4 Omni to generate nuanced financial sentiment examples, this study has provided empirical evidence that LLMs can effectively augment existing datasets, thereby addressing data scarcity and improving model robustness. This use of LLMs for targeted augmentation of BERT like models in financial sentiment analysis is one of the first of its kind to the best of our knowledge, extending the utility of LLMs in practical financial applications.
2. **Improvements in Model Performance**: The augmented data was crucial in improving the accuracy, precision, recall, and F1 scores of the FinBERT model, which was further distilled to develop TinyFinBERT. The performance gains achieved underscore the potential of combining LLM data augmentation with sophisticated distillation techniques to enhance smaller models' capabilities, making them competitive with larger, more resource-intensive models.
3. **Development of New Methodologies**: This research has refined knowledge distillation methodologies, particularly through their adaptation and application within the context of financial NLP. While the structured approach to distillation, including the strategic use of synthetic data, is not entirely new, this study showcases its successful application in financial NLP tasks. This methodology can be leveraged in other domains where model size and performance are critical considerations, demonstrating its broad applicability and potential for cross-domain utility.

4. **Implications for NLP and Financial Analytics**: The successful application of these techniques has broader implications for the field of NLP, particularly in how synthetic data can be utilized to overcome the limitations of available training datasets. Additionally, the ability of smaller models like TinyFinBERT to perform at par with larger models in the financial domain with the help of LLM Data Augmentation opens new avenues for deploying advanced NLP applications in resource-constrained environments, such as mobile devices or in real-time trading systems, where quick processing is essential.

Overall, this research fills a gap in existing literature by demonstrating the successful application of LLM data augmentation techniques and their impact on model performance in financial sentiment analysis. It is hoped that these contributions will influence further research and development in the innovative use of LLM data augmentation for NLP applications in financial markets.

## 6.4 Future Recommendations

This thesis has demonstrated the potential of LLM data augmentation to significantly enhance the performance of financial sentiment analysis models like FinBERT and TinyBERT. Building on these findings, several areas can be further explored to maximize the utility and efficiency of such an approach in natural language processing:

1. **Expansion to Other NLP Domains:** Testing the methodologies used in this thesis across different NLP applications, such as healthcare or legal sectors, could help in identifying domain-specific challenges and opportunities for applying LLM data augmentation.

2. **In-Depth Hyperparameter Optimization:** While this research validated the effectiveness of LLM data augmentation, it did not focus extensively on hyperparameter tuning. Future studies could delve into detailed analyses of hyperparameter settings to optimize model performance further, which is crucial for achieving the highest efficiency and accuracy in machine learning models.

3. **Optimal Data Utilization:** The ratio of augmented to original data used in training was not exhaustively explored. Investigating the ideal balance between these data sources could lead to insights on maximizing training effectiveness and enhancing the model's accuracy and generalization capabilities. This exploration is essential for leveraging the full potential of synthetic data in training robust models.

4. **Computational Efficiency:** Developing strategies to reduce computational demands while maintaining high accuracy is paramount, especially for applications requiring real-time analysis. Future research can aim to innovate more computationally efficient algorithms or improve existing methodologies through the innovative use of advanced LLM models.

5. **Impact of Varied Synthetic Data Types:** Exploring how different types of synthetic data impact model learning and performance could uncover new ways to enhance model robustness and error handling.

By addressing these areas, future research can extend the benefits of advanced NLP techniques, particularly in how machine learning models enhanced with LLM data augmentation can be optimized and applied effectively in various real-world scenarios.

## 6.5   Summary

This chapter has consolidated the major findings and implications of the research conducted on enhancing financial sentiment analysis models through LLM data augmentation and knowledge distillation techniques. The discussions have highlighted how the integration of synthetic data into the training process of FinBERT, subsequently termed Augmented FinBERT, significantly improved model performance across various datasets. The distillation of this enhanced model into a smaller, efficient version, TinyFinBERT, demonstrated the practical application of knowledge distillation strategies, maintaining high accuracy with reduced model size.

Key contributions of this research have been underscored, emphasizing the novel application of LLM data augmentation in the field of financial sentiment analysis. This study not only advanced the technical capabilities of sentiment analysis models but also contributed to the broader NLP community by introducing refined methodologies that can be adapted across other domains.

Looking forward, several avenues for further research were identified, including the exploration of optimal hyperparameter settings and the balance of augmented to original data in training processes. These recommendations aim to guide future efforts in enhancing the efficacy and efficiency of NLP models, ensuring they remain adaptable and robust in various application scenarios.